\documentclass[manuscript,screen]{acmart}

\usepackage{mathtools}
\usepackage{array}
\usepackage{booktabs}
\usepackage{graphicx}
\graphicspath{{./figures/shared/}}
\usepackage{makecell}
\usepackage{multirow}
\usepackage{placeins}

\acmJournal{TOIS}
\setcopyright{none}
\acmDOI{}
\begin{document}

\title[EEG-MedRAG]{EEG-MedRAG: Enhancing EEG-based Clinical Decision-Making via Hierarchical Hypergraph Retrieval-Augmented Generation}
\thanks{A preprint of this work is available at \href{https://arxiv.org/abs/2508.13735}{arXiv:2508.13735}.}
\thanks{This study was supported by the Advanced Materials-National Science and Technology Major Project (Grant No.~2024ZD0607700), the Liaoning Province Key Scientific Research Project (2026JH1/11700023), the Natural Science Foundation of Liaoning Province (2026-MS-044), and the Guangdong Basic and Applied Basic Research Foundation (2025A1515011602).}

\author{Lu Meng}
\authornote{These authors contributed equally to this work.}
\email{menglu@ise.neu.edu.cn}
\affiliation{%
  \institution{College of Information Science and Engineering, Northeastern University}
  \city{Shenyang}
\country{China}}
\affiliation{%
  \institution{Foshan Graduate School of Innovation, Northeastern University}
  \city{Foshan}
\country{China}}

\author{Zan Li}
\authornotemark[1]
\email{lizan@mails.neu.edu.cn}
\affiliation{%
  \institution{College of Information Science and Engineering, Northeastern University}
  \city{Shenyang}
\country{China}}

\author{Yi Wang}
\email{2210372@stu.neu.edu.cn}
\affiliation{%
  \institution{College of Information Science and Engineering, Northeastern University}
  \city{Shenyang}
\country{China}}

\author{Yi Ding}
\email{ding.yi@ntu.edu.sg}
\affiliation{%
  \institution{College of Computing and Data Science, Nanyang Technological University}
  \city{Singapore}
\country{Singapore}}

\author{Ziyue Zhu}
\email{zuzuzzy@hotmail.com}
\affiliation{%
  \institution{College of Computing and Data Science, Nanyang Technological University}
  \city{Singapore}
\country{Singapore}}

\author{Fang Li}
\email{asfli@ntu.edu.sg}
\affiliation{%
  \institution{College of Computing and Data Science, Nanyang Technological University}
  \city{Singapore}
\country{Singapore}}

\author{Ziyu Jia}
\email{jia.ziyu@outlook.com}
\affiliation{%
  \institution{Institute of Automation, Chinese Academy of Sciences}
  \city{Beijing}
\country{China}}

\author{Anh Tuan Luu}
\email{anhtuan.luu@ntu.edu.sg}
\affiliation{%
  \institution{College of Computing and Data Science, Nanyang Technological University}
  \city{Singapore}
\country{Singapore}}

\author{Xinliang Zhou}
\email{xinliang001@e.ntu.edu.sg}
\affiliation{%
  \institution{College of Computing and Data Science, Nanyang Technological University}
  \city{Singapore}
\country{Singapore}}

\author{Qingsong Wen}
\email{qingsongedu@gmail.com}
\affiliation{%
  \institution{Squirrel Ai Learning}
  \city{Bellevue}
  \state{WA}
\country{USA}}

\author{Haoran Luo}
\authornote{Corresponding author: Haoran Luo.}
\email{haoran.luo@ntu.edu.sg}
\affiliation{%
  \institution{College of Computing and Data Science, Nanyang Technological University}
  \city{Singapore}
\country{Singapore}}

\renewcommand{\shortauthors}{Meng et al.}

\begin{abstract}
  With the widespread application of electroencephalography (EEG) in neuroscience and clinical practice, efficiently retrieving and semantically interpreting large-scale, multi-source, heterogeneous EEG data has become a pressing challenge. EEG-MedRAG addresses this challenge through a three-layer hypergraph-based retrieval-augmented generation framework that unifies EEG domain knowledge, individual patient cases, and large-scale EEG recordings into a traversable n-ary relational hypergraph, enabling joint semantic-temporal retrieval and evidence-grounded clinical reasoning.
  The accompanying benchmark is the first to pair EEG questions about the same clinical fact across five user roles. Its 700 questions span seven clinical domains and support direct comparisons of answer consistency across roles.
  Across three LLM backbones, EEG-MedRAG outperforms all four baselines in both overall answer correctness and token-level F1. It achieves up to 86.14\% answer correctness on the full benchmark and 97.32\% on non-EEG questions, with 97.41\% evidence coverage on knowledge questions. In EEG retrieval, EEG-MedRAG improves top-five label agreement by 8.58 percentage points over the time-series baseline.
  The code is available at \url{https://github.com/li-zan/EEG-MedRAG}.
\end{abstract}



\maketitle

\section{Introduction}
Electroencephalography (EEG) provides non-invasive, high-temporal-resolution access to brain dynamics and is foundational to epilepsy diagnosis, sleep disorder analysis, and the monitoring of neurodegenerative conditions \cite{alsharabi2023eeg,uyanik2025automated,chen2023automated,barnes2022detection,wang2022eeg,zhang2023applied}. With the proliferation of EEG devices and the expansion of multi-center data collection \cite{casson2019wearable,michel2012towards}, large-scale, multi-source EEG corpora are rapidly accumulating. This growth creates unprecedented opportunities to characterize brain dynamics at fine temporal granularity and to develop more robust, intelligent interpretation methods \cite{Yi2022WTCS,Zhang2020Intention}. Despite this centrality, efficiently retrieving and interpreting large-scale, multi-source EEG resources remains difficult \cite{singh2023trends,roy2019deep,bustios2023incorporating,craik2019deep}. In practice, clinical interpretation is not simply the classification of static clips but rather a continuous and context-bound reasoning process that unfolds across the signal and the clinical encounter. Technologist annotations, stimulus protocols and sleep-stage transitions, prior seizures and medications, comorbidities, and the diagnostic intent jointly shape conclusions \cite{Phyo2023TransSleep,Li2022Seizure}. At the same time, real-world EEG exhibits pronounced heterogeneity, so models that perform well under controlled conditions often degrade when deployed \cite{Li2020MSTL,Qi2021STECS}. Research on physiological signal analysis and longitudinal health modeling highlights the complementary roles of multimodal observations, temporal context, and structured medical knowledge \cite{Zheng2019EmotionMeter,Phyo2023TransSleep,Lu2023Sherbet}. Real clinical use further requires joint reasoning over domain knowledge, patient-specific records, and session- or stream-level EEG signals so that generated assessments are precise and context-aware \cite{zhao2025medrag,ren2024healthcare,cyril2025almanac}. These needs expose a gap between the availability of heterogeneous EEG corpora and the ability of current systems to integrate them into dependable clinical decision support.

\begin{figure}
  \centering
  \includegraphics[width=0.70\linewidth]{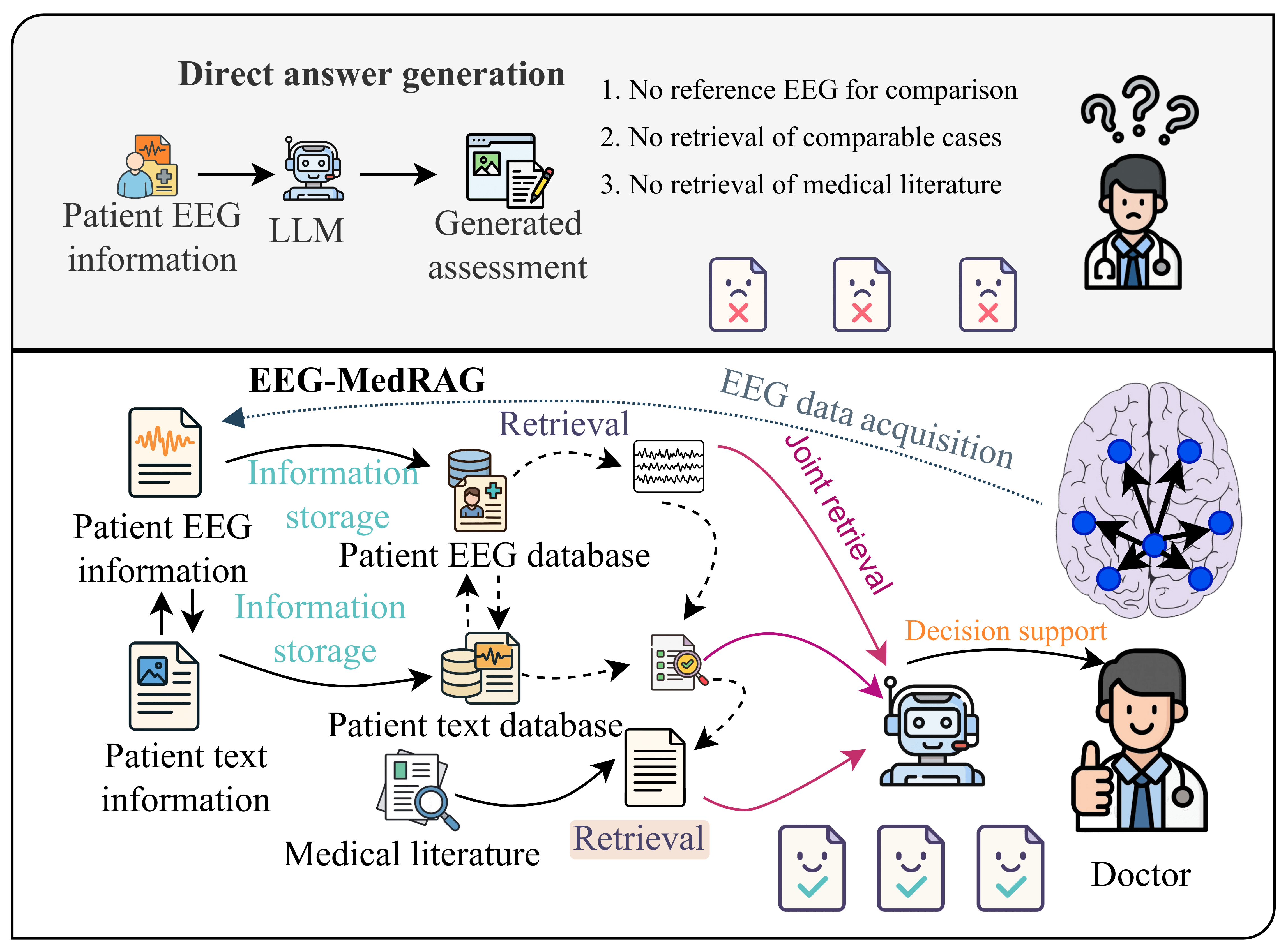}
  \caption{Two workflows for EEG assessment. Direct generation relies on the patient information supplied with the query (top). EEG-MedRAG also draws on reference EEG recordings, comparable cases, and medical literature to inform the assessment (bottom).}
  \Description{Two stacked panels compare clinical workflows. The top panel shows direct answer generation from patient EEG information through an LLM without reference EEG, comparable cases, or medical literature, leaving a physician uncertain. The bottom panel shows EEG-MedRAG storing patient EEG and text records alongside medical literature for joint retrieval, providing grounded decision support to the physician.}
  \label{fig:eeg-assessment-workflows}
\end{figure}

Retrieval-Augmented Generation (RAG) has recently improved knowledge-intensive generation by coupling LLMs with external retrieval \cite{wu2024medical}. However, standard chunk-based RAG neglects the \emph{structured relationships} among medical entities that guide clinical reasoning \cite{lewis2020retrieval,yasunaga2022linkbert}. Graph-based RAG narrows this gap by retrieving over entity-relation structures and often yields better precision \cite{wu2024medical,peng2024graph,han2024graphrag}. Yet most formulations still encode predominantly binary relations and thus fall short of capturing the $n$-ary interactions ubiquitous in EEG practice, such as waveform-to-symptom-to-diagnosis chains that link physiological patterns to phenotypes and downstream decisions \cite{luo2025hypergraphrag,luo2023hahe}. Temporal-aware variants alleviate part of the problem on the signal side, but they typically lack principled integration with structured domain knowledge and patient context \cite{yang2024timerag}. The resulting mismatch leads to fragmented representations, brittle retrieval, and diminished reliability in clinical settings.

In this work, we formalize three obstacles that must be addressed to close this gap. First, limitations in modeling $n$-ary clinical knowledge cause fragmented representations and reduce diagnostic fidelity. Second, existing pipelines struggle to jointly retrieve temporally aligned EEG segments together with semantically matched patient records and medical concepts, which undermines faithful grounding of LLM reasoning. Third, existing EEG QA benchmarks lack a common basis for comparing how well systems answer questions across clinical domains and user roles. Together, these gaps restrict the reliability of existing RAG pipelines in clinical practice. Figure~\ref{fig:eeg-assessment-workflows} contrasts direct LLM generation with the proposed framework.

\begin{figure*}
  \centering
  \includegraphics[width=\textwidth]{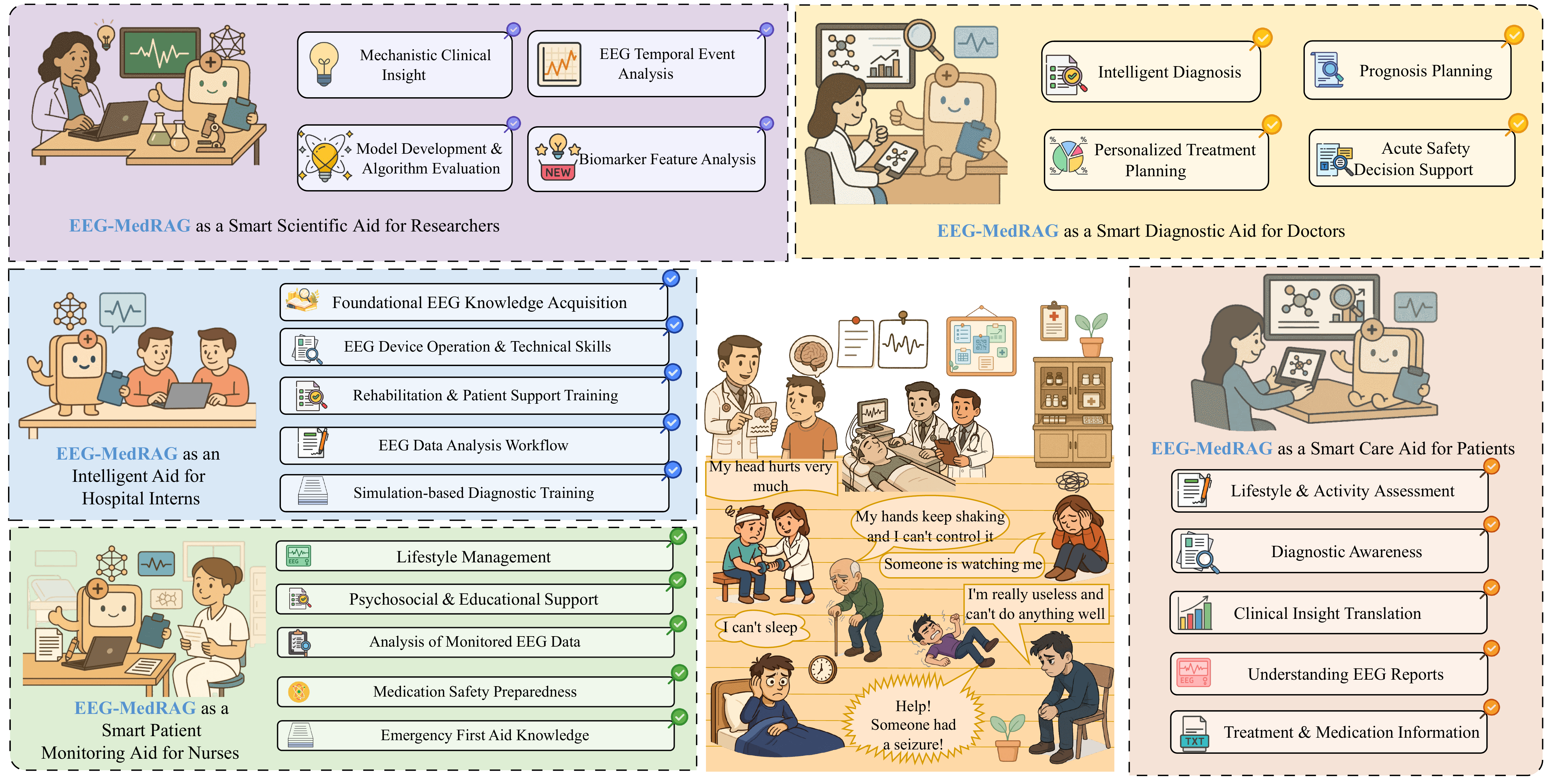}
  \caption{Illustrative EEG-related information needs of physicians, clinical interns, nurses, patients, and researchers in the EEG-MedRAG benchmark, which spans seven clinical domains.}
  \Description{Five panels surround a clinical scene. Each pairs a user role with examples of information needs, including EEG analysis for researchers, diagnostic assessment for physicians, and explanations of EEG reports for patients.}
  \label{fig:clinical-roles-and-tasks}
\end{figure*}

To address these challenges, we propose EEG-MedRAG, a three-layer hypergraph-based RAG framework for EEG clinical decision making. The framework constructs a structured hypergraph that integrates domain knowledge, patient cases, and large-scale EEG data into a traversable $n$-ary substrate, enabling explicit representation of multi-entity chains. On top of this structure, a joint semantic-temporal retrieval strategy aligns EEG segments with patient history and medical concepts to strengthen grounding and context integration (Figure~\ref{fig:eeg-medrag-framework}). We also introduce the first clinical EEG question answering benchmark to pair questions about the same underlying fact across five user roles in each of seven clinical domains (Figure~\ref{fig:clinical-roles-and-tasks}). Together, these components provide a unified route to encode, retrieve, and utilize heterogeneous EEG evidence for clinically meaningful generation.

We evaluate EEG-MedRAG across seven public clinical EEG cohorts, a shared medical knowledge corpus, and three generation backbones. Across all settings, EEG-MedRAG achieves consistent gains in answer correctness (81.29\%--86.14\%), token-level F1, evidence coverage (97.41\%), and context relevance over NaiveGeneration, StandardRAG, HyperGraphRAG \cite{luo2025hypergraphrag,luo2023reasoning}, and TimeRAG \cite{yang2024timerag}. Leave-one-out and single-layer ablations confirm the complementary value of the knowledge hypergraph, case hypergraph, and EEG repository, while full-repository banded PAA--DTW improves waveform-neighbor concordance over clustered time-series retrieval (49.29\% versus 40.71\% LA@5). Structured cross-layer grounding further achieves up to 97.32\% accuracy on non-EEG clinical workflows and maintains consistent performance across all five clinical roles. The remainder of this paper is organized as follows: Section~\ref{sec:related} reviews related work, Sections~\ref{sec:problem}--\ref{sec:method} formalize the problem and present EEG-MedRAG, Section~\ref{sec:benchmark} introduces the benchmark, Sections~\ref{sec:setup}--\ref{sec:results} report the experimental setup and results, and Sections~\ref{sec:discussion}--\ref{sec:conclusion} provide further discussion and conclude.

\section{Related Work}
\label{sec:related}

\subsection{Retrieval-Augmented Generation for Medical Question Answering}

Medical language models have been developed through domain-specific instruction tuning and the incorporation of patient context. MedAlpaca explores instruction tuning for medical question answering, while work on medical assistant personalization integrates short- and long-term memory to account for users' backgrounds and preferences \cite{han2023medalpaca,zhang2023llm}. Retrieval-augmented generation provides a complementary approach by conditioning responses on external evidence. The original RAG framework combines passage retrieval with language generation, and Self-RAG incorporates adaptive retrieval and self-reflection \cite{lewis2020retrieval,asai2024self}. In medical applications, TC-RAG introduces state-aware control of retrieval and reasoning, while RetCare draws on medical literature to support the interpretation of clinical model predictions \cite{jiang2024tcrag,wang2024retcare}.

Another line of research extends language models to time-series data. Time-LLM reprograms time-series inputs for forecasting with pretrained language models, while TimeRAG retrieves similar historical sequences to inform forecasts \cite{jin2023timellm,yang2024timerag}. Their focus on forecasting leaves a broader retrieval problem open: how to connect temporal observations with the knowledge and clinical context needed to answer medical questions. In EEG analysis, waveform similarity alone does not establish clinical relevance, and relevant textual passages may not capture the physiological patterns underlying a question. Treating these sources independently fragments the evidence needed for clinical reasoning. The challenge is therefore to retrieve across medical knowledge, patient histories, and EEG recordings while preserving the relationships that make their joint interpretation meaningful. Addressing this challenge broadens retrieval-augmented medical question answering to encompass both semantic and physiological evidence.

\subsection{Graph and Hypergraph-based Retrieval}

Structured knowledge provides relational context for language generation and reasoning. KG-BART incorporates relations among concepts into commonsense generation, while DyKgChat and knowledge graph-augmented dialogue models investigate how structured knowledge can guide conversational responses \cite{liu2021kg,tuan2019dykgchat,kang2023knowledge}. For knowledge graph question answering, ReasoningLM integrates subgraph structure into language modeling, and Reasoning on Graphs uses relation-based plans to retrieve reasoning paths \cite{jiang2023reasoninglm,luo2023reasoning}. In the medical domain, knowledge graph-assisted dialogue generation connects conversational context with medical entities and their relations \cite{varshney2023knowledge}. Together, these approaches establish relational structure as a resource for selecting and organizing evidence beyond isolated text passages.

The representation of relational evidence is itself consequential. Decomposing a fact involving several entities into separate binary relations can obscure the shared context that gives the fact its meaning. HyperGraphRAG represents n-ary facts as hyperedges and combines entity and hyperedge retrieval to construct the generation context \cite{luo2025hypergraphrag}. This preserves associations among multiple entities within a unified fact representation. For EEG question answering, however, preserving relational facts addresses only part of the retrieval problem: their clinical relevance also depends on the patient context and physiological observations to which they apply. Retrieving a relevant knowledge structure without linking it to these observations leaves a gap between general medical knowledge and the evidence needed to reason about a particular case.

EEG-MedRAG addresses structured retrieval for clinical questions whose evidence spans medical literature, clinical records, and EEG recordings. Clinical records provide the context for interpreting medical knowledge, while waveform similarity connects a query to reference recordings and their associated cases. Linking these sources brings relational knowledge and physiological evidence into a common retrieval framework, supporting clinical question answering through the joint use of semantic relevance, case context, and temporal similarity.

\subsection{EEG Analysis and Clinical Decision Support}

Deep learning and structured signal analysis have advanced automated EEG interpretation across clinical tasks such as seizure detection, sleep staging, and neurodegenerative assessment~\cite{roy2019deep,craik2019deep,singh2023trends,Li2022Seizure,Phyo2023TransSleep,wang2022eeg}. Subsequent studies address cross-subject variability, channel selection, and multimodal physiological cues to improve generalization across recordings~\cite{Li2020MSTL,Qi2021STECS,Zheng2019EmotionMeter,Lu2023Sherbet}. Yet these models are typically developed as task-specific classifiers for a fixed label space and electrode montage; they output narrow predictions rather than evidence-grounded answers that interpret physiological findings in light of clinical guidelines and patient history.

In parallel, retrieval-augmented clinical language models have been explored to support medical question answering, guideline consultation, and longitudinal care workflows~\cite{cyril2025almanac,ren2024healthcare,zhao2025medrag,wu2024medical}. However, existing medical QA systems and benchmarks operate primarily over textual literature or structured health records and rarely incorporate multichannel EEG recordings directly into retrieval. Moreover, clinical evaluations seldom test whether a system remains factually consistent when the same underlying finding is queried across distinct clinical roles, leaving a gap between waveform modeling and role-aware clinical decision support.

Our work bridges these directions along two dimensions. Methodologically, EEG-MedRAG combines domain-specific multichannel waveform matching with case-level and knowledge-level hypergraph retrieval so that physiological similarities are interpreted through comparable patient records and $n$-ary medical knowledge. Empirically, our cross-disease, cross-role benchmark pairs seven public clinical EEG cohorts with five stakeholder perspectives under strict subject-level partitioning, enabling systematic evaluation of multimodal clinical EEG question answering.

\section{Problem Formulation}
\label{sec:problem}

\subsection{Task Definition}

We study retrieval-augmented question answering for EEG-related clinical questions posed from the perspectives of physicians, nurses, clinical interns, patients, and researchers. Questions posed from different perspectives may concern the same clinical fact. The task is to identify evidence relevant to each question and use it to provide an accurate answer.

A query $q$ specifies the question and the role from which it is posed. General questions about EEG need no patient record. For a case-specific question, the query also provides the clinical observations available for that question and, when applicable, an EEG segment $x_q$. A finding that the question asks the system to infer is withheld from the accompanying case information. We denote the question together with its accompanying clinical text by $t_q$. Medical knowledge and observations from reference cases can inform the interpretation, while conclusions about the patient must be grounded in the observations available for that patient.

\subsection{Retrieval Setting}

Evidence comes from three sources: a shared collection of EEG and clinical literature, reference case records, and EEG recordings from reference subjects. The literature is available for all questions. For case-specific questions, reference cases can be retrieved across clinical domains. When a query includes an EEG segment, waveform comparisons are restricted to reference segments from the same domain. Reference and test subjects are disjoint: all cases and recordings from a given subject belong to the same partition.

We denote the objects used in retrieval by
\begin{equation*}
  U = V \cup E^{H} \cup P \cup X,
\end{equation*}
where $V$ is the set of medical entities, $E^{H}$ the set of knowledge hyperedges expressing facts that relate these entities, $P$ the set of reference cases, and $X$ the set of reference EEG segments. Entities and facts are linked to their source passages, and EEG segments to their corresponding cases. These associations preserve the context needed to interpret a retrieved fact or a waveform match.

For an object $u$ with a textual representation, let $f_{\mathrm{text}}(u)$ denote its text embedding. We measure textual similarity by
\begin{equation}
  \label{eq:sim}
  s(u,v) = \cos\!\left(f_{\mathrm{text}}(u), f_{\mathrm{text}}(v)\right) =
  \frac{\langle f_{\mathrm{text}}(u), f_{\mathrm{text}}(v)\rangle}
  {\lVert f_{\mathrm{text}}(u)\rVert_2\,\lVert f_{\mathrm{text}}(v)\rVert_2}.
\end{equation}
EEG retrieval compares multichannel waveforms over time. When a reference segment matches the query waveform, its associated case information, recording conditions, and any events annotated during that interval provide evidence for answering the question.

\section{EEG-MedRAG Framework}
\label{sec:method}

\subsection{Framework Overview}

\begin{figure*}
  \centering
  \includegraphics[width=\textwidth]{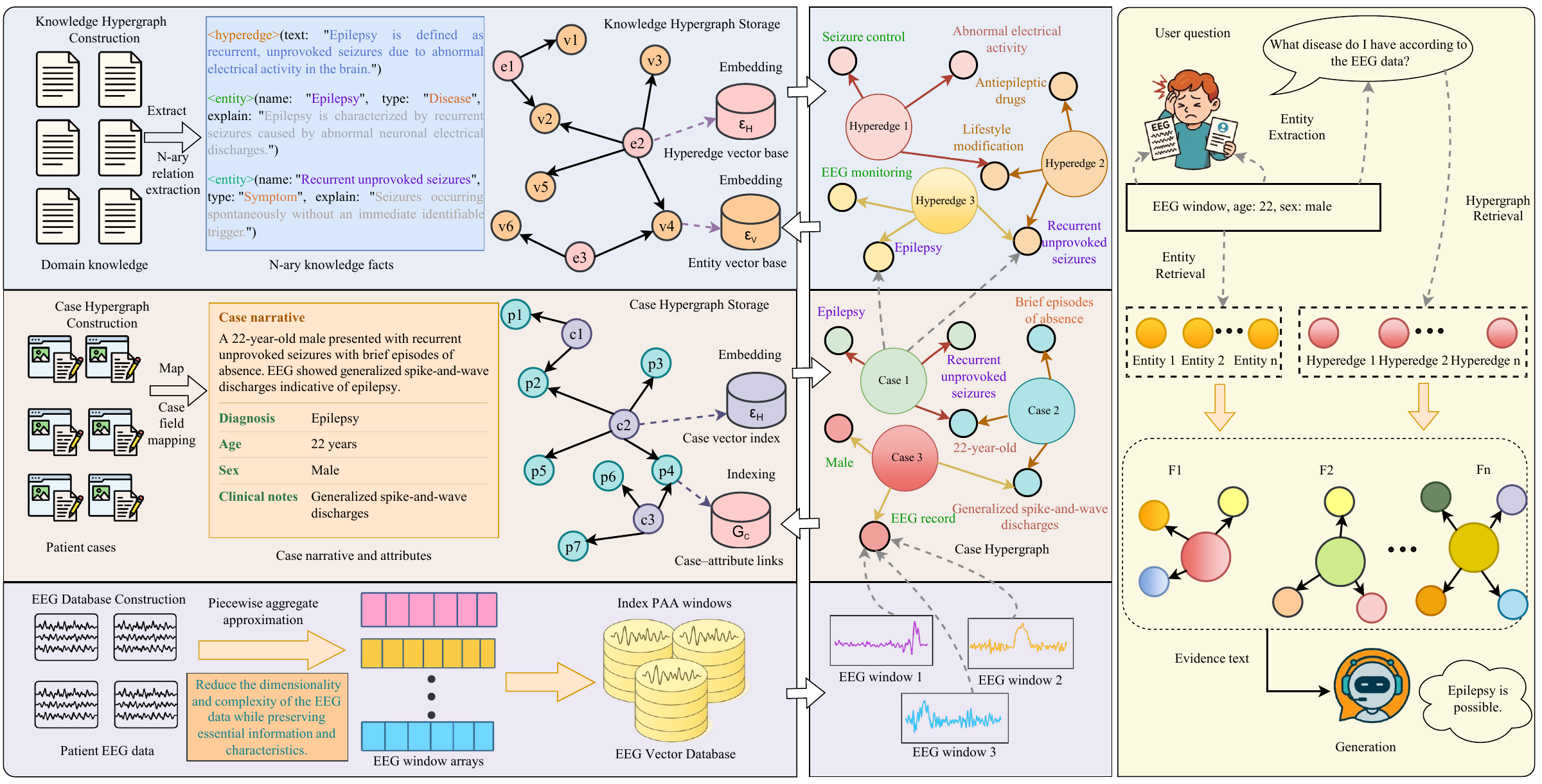}
  \caption{Overview of EEG-MedRAG, which integrates knowledge and case hypergraphs with a repository of EEG windows to retrieve complementary evidence for clinical question answering.}
  \Description{On the left, medical literature and patient records are organized into separate hypergraphs, and EEG recordings are indexed as windows. The center shows how a reference case links EEG windows to medical concepts. On the right, a patient's question, age, sex, and EEG window guide retrieval. The model uses the retrieved evidence to identify epilepsy as a possibility.}
  \label{fig:eeg-medrag-framework}

\end{figure*}

EEG-MedRAG organizes medical knowledge, clinical cases, and EEG recordings within a three-layer hypergraph-based framework (Figure~\ref{fig:eeg-medrag-framework}). The framework combines semantic and temporal retrieval with cross-layer evidence integration to support clinical reasoning and answer generation.

\subsection{Hierarchical Hypergraph Construction}
\label{sec:construction}

\paragraph{Knowledge Hypergraph (KHG)}
The first layer, KHG, encodes medical knowledge from clinical guidelines, interpretation standards, and scientific articles as $n$-ary facts, each comprising a textual statement and its participating entities. Source documents are partitioned into a collection of text chunks $D$. For each chunk $d\in D$, a language model $\pi$, guided by an extraction prompt $p_{\text{ext}}$, jointly extracts factual statements and their participating entities:
\begin{equation*}
  \mathcal{F}_d = \{(e_i^H,V_i^H)\}_{i=1}^{k_d} \sim \pi(\,\cdot\mid d,p_{\text{ext}}),
\end{equation*}
where $k_d$ is the number of extracted facts, $e_i^H$ denotes a knowledge hyperedge carrying a natural-language statement, and $V_i^H$ denotes the set of entities participating in that statement. Each entity is represented by its name, type, and contextual description. The number of participating entities may vary across facts, and an entity may participate in multiple facts.

The extracted records form the knowledge hypergraph $G_{\mathrm{KHG}}=(V,E^H)$. We store it as an incidence bipartite graph with two disjoint node sets: one for entities and one for knowledge hyperedges. Each incidence link connects a hyperedge node to an entity participating in the corresponding fact:
\begin{equation*}
  G_{\mathrm{BKG}}=(V\sqcup E^H,I),\qquad
  I=\{(e^H,v)\mid e^H\in E^H,\ v\in V(e^H)\},
\end{equation*}
where $I$ denotes the incidence-link set and $V(e^H)$ denotes the entities associated with hyperedge $e^H$, with $V(e_i^H)=V_i^H$. This structure supports traversal from a fact to its participating entities and from an entity to the facts in which it appears.

Entities and hyperedges are embedded using the shared text encoder $f_{\mathrm{text}}$ and stored in separate vector indexes. Entity embeddings encode names and descriptions, while hyperedge embeddings encode factual statements. The records retain source-chunk identifiers for provenance.

\paragraph{Case Hypergraph (CHG)}
The second layer, CHG, represents each clinical case $p\in P$ as a hyperedge connecting its recorded medical attributes. These include available demographics, diagnoses, clinical assessments, clinical notes, and recording conditions.

Attributes are represented as field--value pairs, retaining the associated assessment metadata where available. We store the resulting CHG as an incidence bipartite graph between case hyperedges and attribute nodes, with identical attribute representations shared across cases.

The recorded medical attributes of each case are rendered as a clinical narrative, which we encode as
\begin{equation}
  \label{eq:case-embedding}
  f_{\mathrm{CHG}}(p)=f_{\mathrm{text}}(t_p),
\end{equation}
where $t_p$ denotes the narrative of case $p$ and $f_{\mathrm{text}}$ is the shared text encoder. The resulting case embeddings are stored in a vector index.

We establish soft links from each case hyperedge to its most similar KHG entities and hyperedges, enabling cross-layer retrieval that connects clinical case evidence with supporting medical knowledge for clinical reasoning. Candidate link weights are given by
\begin{equation*}
  w_{p\to u}=s(p,u),\qquad u\in V\cup E^H,
\end{equation*}
where $s(p,u)$ is the cosine similarity between the embedding of the case narrative $t_p$ and that of knowledge node $u$, as defined in Eq.~\eqref{eq:sim}. For each case $p$, we retain up to $k_{\mathrm{soft}}$ highest-scoring knowledge nodes whose similarity exceeds a threshold as its cross-layer soft links, which later seed neighborhood expansion in Eq.~\eqref{eq:expand}.

\paragraph{EEG Vector Database (EVD)}
The third layer represents each EEG window as a channel-by-time matrix. Given a window $x\in\mathbb{R}^{C\times T}$, we apply piecewise aggregate approximation (PAA)~\cite{keogh2001dimensionality} independently to each channel:
\begin{equation}
  \label{eq:eeg-window-representation}
  Z_x=f_{\mathrm{EEG}}(x)=
  \begin{bmatrix}
    \mathrm{PAA}(x_1,n)\\
    \vdots\\
    \mathrm{PAA}(x_C,n)
  \end{bmatrix}
  \in\mathbb{R}^{C\times n},
\end{equation}
where $C$ is the number of channels, $T$ is the number of samples per channel, and $n$ is the number of PAA segments per channel. Each row contains the PAA values of one channel, and each column corresponds to the same time interval across channels.

The representation retains the channel order and the temporal positions of the PAA segments for multichannel similarity retrieval. Each EEG segment is independently encoded by \(f_{\mathrm{EEG}}\) and stored in the EVD, with identifiers linking it to its source recording and corresponding case.

The input EEG windows follow domain-specific channel and reference conventions and undergo band-pass filtering, resampling, and technical quality screening during dataset preparation.

For each channel, the PAA values are computed as follows:
\begin{equation}
  \label{eq:paa}
  \begin{aligned}
    m &= \left\lfloor \frac{T}{n} \right\rfloor,\\
    \mathrm{PAA}(x_c,n)_j &= \frac{1}{m}\sum_{t=(j-1)m+1}^{jm}x_c(t),\qquad j=1,\dots,n ,
  \end{aligned}
\end{equation}
where $m$ is the number of samples per PAA segment, $\lfloor\cdot\rfloor$ denotes the floor operator, $x_c(t)$ is the preprocessed voltage of channel $c$ at sample $t$, and $\mathrm{PAA}(x_c,n)_j$ is the mean over the $j$th segment. The $n$ non-overlapping segments cover the first $mn$ samples.

\subsection{Semantic-Temporal Retrieval}
\label{sec:retrieval}
EEG-MedRAG retrieves medical knowledge, reference cases, and EEG windows as complementary evidence for clinical reasoning. The question and accompanying clinical context form the textual query $t_q$, while EEG retrieval uses the available waveform input.

\paragraph{Knowledge Retrieval}
We use a language model to extract medical entity mentions from $t_q$ and concatenate them into an entity query $t_q^V$. Entity retrieval uses $t_q^V$, while hyperedge retrieval uses the complete query $t_q$:
\begin{equation*}
  \begin{aligned}
    R_V(t_q)
    &=\underset{v\in V}{\operatorname{Top}_{k_V}}
    \cos\!\left(f_{\mathrm{text}}(t_q^V),f_{\mathrm{text}}(v)\right),\\
    \mathcal C_H(t_q)
    &=\underset{e^H\in E^H}{\operatorname{Top}_{k_{\mathrm{cand}}}}
    \cos\!\left(f_{\mathrm{text}}(t_q),f_{\mathrm{text}}(e^H)\right),
  \end{aligned}
\end{equation*}
where $R_V(t_q)$ contains up to $k_V$ retrieved entities and $\mathcal C_H(t_q)$ contains up to $k_{\mathrm{cand}}$ candidate hyperedges.

To reduce redundancy among retrieved facts, we select hyperedges from $\mathcal C_H(t_q)$ using greedy maximal marginal relevance (MMR)~\cite{carbonell1998mmr}. Starting with $R_H^{(0)}=\varnothing$, step $j$ selects
\begin{equation*}
  \begin{aligned}
    e_j^H
    ={}&\underset{e^H\in\mathcal C_H(t_q)\setminus R_H^{(j-1)}}
    {\operatorname{arg\,max}}
    \Bigg[
      \lambda\cos\!\left(f_{\mathrm{text}}(t_q),f_{\mathrm{text}}(e^H)\right)\\
      &\qquad-(1-\lambda)
      \max_{e'\in R_H^{(j-1)}}
      \cos\!\left(f_{\mathrm{text}}(e^H),f_{\mathrm{text}}(e')\right)
    \Bigg],
  \end{aligned}
\end{equation*}
where $\lambda\in[0,1]$ balances query relevance and redundancy with previously selected hyperedges. The redundancy term is zero when $R_H^{(j-1)}$ is empty. We update $R_H^{(j)}=R_H^{(j-1)}\cup\{e_j^H\}$ until $k_H$ hyperedges have been selected or the candidate pool is exhausted. The retrieved entities $R_V(t_q)$ and the final hyperedge set $R_H(t_q)$ serve as seeds for cross-layer expansion.

\paragraph{Case Retrieval}
Given the textual query $t_q$, we retrieve reference case hyperedges $p\in P$ through narrative similarity and attribute matching in CHG. Narrative retrieval yields the candidate list
\begin{equation*}
  \mathcal C_P^{\mathrm{text}}(t_q)
  =\underset{p\in P}{\operatorname{Top}_{k_P}}
  \cos\!\left(f_{\mathrm{text}}(t_q),f_{\mathrm{CHG}}(p)\right),
\end{equation*}
where $k_P$ is the case retrieval cutoff and $f_{\mathrm{CHG}}(p)$ is the case-narrative embedding defined in Eq.~\eqref{eq:case-embedding}.

The attribute route retrieves case hyperedges through their incident medical attributes. Let $\mathcal A_q(p)$ denote the distinct query attributes matched by case $p$, and let $\mathcal A_q^{\mathrm{clin}}(p)\subseteq\mathcal A_q(p)$ contain matches involving diagnoses, clinical assessments, medication or sleep states, and visit stages. Matches require agreement on attribute identity and value, together with any supplied assessment qualifiers. The attribute candidates are
\begin{equation*}
  \mathcal C_P^{\mathrm{attr}}(t_q)
  =\underset{\substack{p\in P\\\mathcal A_q^{\mathrm{clin}}(p)\ne\varnothing}}
  {\operatorname{Top}_{k_P}^{\mathrm{lex}}}
  \left(
    |\mathcal A_q(p)|,\,
    \cos\!\left(f_{\mathrm{text}}(t_q),f_{\mathrm{CHG}}(p)\right)
  \right),
\end{equation*}
where $\operatorname{Top}_{k_P}^{\mathrm{lex}}$ orders the pairs lexicographically in descending order, prioritizing the number of matched attributes and using narrative similarity to resolve ties.

We combine the two candidate rankings using reciprocal rank fusion (RRF)~\cite{cormack2009reciprocal}:
\begin{equation*}
  \begin{aligned}
    s_{\mathrm{RRF}}(p)
    &=\frac{1}{c_{\mathrm{RRF}}+\operatorname{rank}_{\mathrm{text}}(p)}
    +\frac{1}{c_{\mathrm{RRF}}+\operatorname{rank}_{\mathrm{attr}}(p)},\\
    R_P(t_q)
    &=\underset{p\in\mathcal C_P^{\mathrm{text}}(t_q)\cup\mathcal C_P^{\mathrm{attr}}(t_q)}
    {\operatorname{Top}_{k_P}}\,s_{\mathrm{RRF}}(p),
  \end{aligned}
\end{equation*}
where $\operatorname{rank}_{\mathrm{text}}(p)$ and $\operatorname{rank}_{\mathrm{attr}}(p)$ are the positions of case $p$ in the two candidate lists, starting from one, with rank $+\infty$ for an absent case. The constant $c_{\mathrm{RRF}}>0$ is the RRF offset. The resulting list $R_P(t_q)$ contains up to $k_P$ retrieved cases, with narrative ranking resolving fused-score ties.

\begin{figure*}
  \centering
  \includegraphics[width=\textwidth]{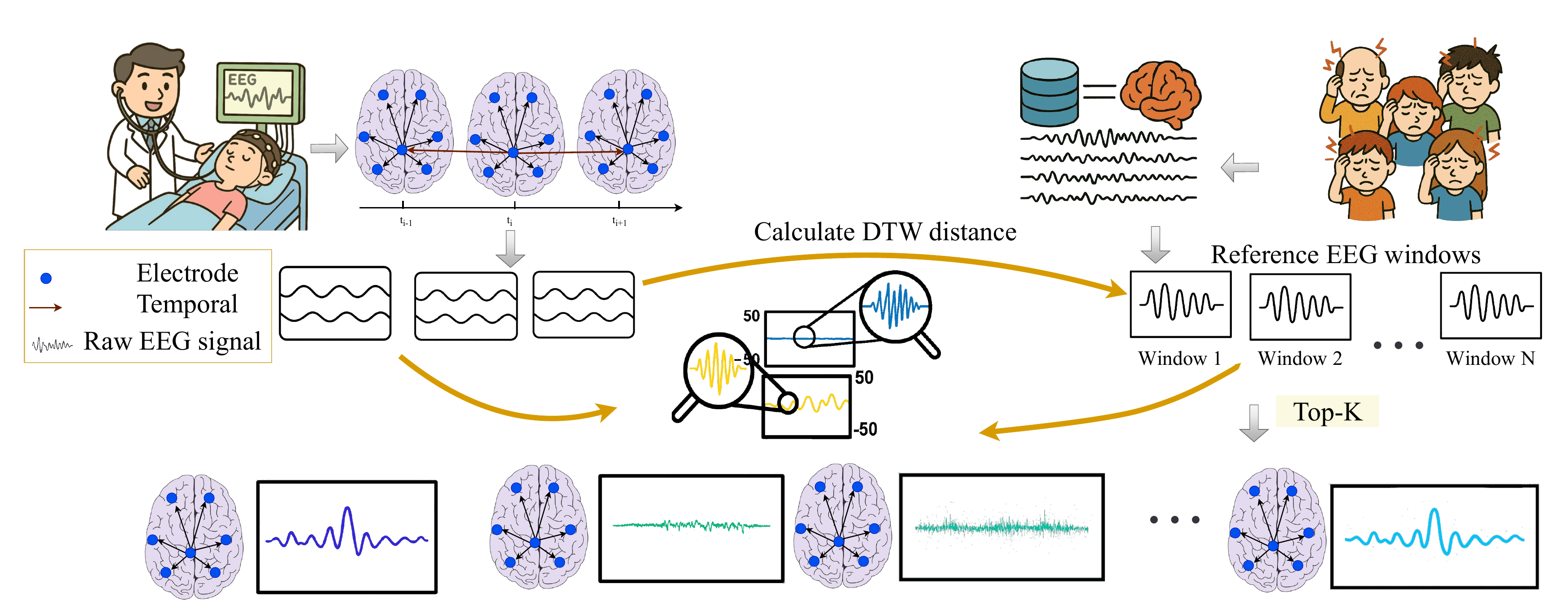}
  \caption{EEG-level retrieval with multichannel PAA representations. Shared-path DTW compares the query window with reference windows and retrieves the top-$K$ matches.}
  \Description{A query EEG window and a database of reference windows feed into a multichannel DTW distance calculation. The reference windows are ranked by distance, and the highest-ranked windows are returned.}
  \label{fig:eeg-window-retrieval}

\end{figure*}

\paragraph{EEG Retrieval}
Given a query EEG window $x_q$, we compute its PAA representation using Eq.~\eqref{eq:eeg-window-representation} and retrieve the top-$K$ reference windows with band-constrained multivariate DTW (Figure~\ref{fig:eeg-window-retrieval}):
\begin{equation*}
  R_{\mathrm{EEG}}(x_q)
  =\underset{x_i\in X_{\mathrm{dom}(q)}}{\operatorname{Top}_K}\,
  s_r\!\big(f_{\mathrm{EEG}}(x_q),\,f_{\mathrm{EEG}}(x_i)\big),
\end{equation*}
where $X_{\mathrm{dom}(q)}\subseteq X$ denotes the reference EEG window repository for the clinical domain of the query and $\operatorname{Top}_K(\cdot)$ returns the $K$ highest-scoring windows under $s_r$. Query and reference windows within a domain share the same ordered channel set and preprocessing protocol. For two PAA matrices $Z,Z'\in\mathbb{R}^{C\times n}$, the similarity is $s_r(Z,Z')=-\mathrm{DTW}_r(Z,Z')$. All channels share one temporal alignment path, with local cost
\begin{equation*}
  d(i,j)=\sum_{c=1}^{C}\left|Z_{c,i}-Z'_{c,j}\right|,
\end{equation*}
where $i$ and $j$ index PAA time positions. The cumulative alignment cost is computed with a Sakoe--Chiba band~\cite{sakoe1978dynamic}:
\begin{equation*}
  D_{i,j}=
  \begin{cases}
    d(i,j)+\min\{D_{i-1,j},\,D_{i,j-1},\,D_{i-1,j-1}\}, & |i-j|\le r,\\[3pt]
    +\infty, & |i-j|>r.
  \end{cases}
\end{equation*}
We initialize $D_{0,0}=0$ and $D_{i,0}=D_{0,j}=+\infty$ for $i,j\ge1$, and set $\mathrm{DTW}_r(Z,Z')=D_{n,n}$. Given a temporal tolerance $\Delta$ in seconds, the radius is
\begin{equation*}
  r=\left\lfloor\frac{\Delta f_s}{m}\right\rfloor,
\end{equation*}
where $f_s$ is the sampling rate and $m/f_s$ is the duration represented by each PAA value, with $m$ defined in Eq.~\eqref{eq:paa}.

\subsection{Cross-Layer Retrieval Fusion}

\paragraph{Cross-Layer Expansion}
Case soft links connect retrieved clinical evidence with related medical knowledge in KHG, with EEG matches linked through their associated cases. We combine the linked knowledge nodes with directly retrieved KHG entities and hyperedges to seed neighborhood expansion on the bipartite knowledge graph:
\begin{equation}
  \label{eq:expand}
  \mathcal{N}_\rho(\mathcal{E}_0)=\{u:\mathrm{dist}_{G_{\mathrm{BKG}}}(u,\mathcal{E}_0)\le \rho\},
\end{equation}
where $\mathcal{E}_0\subseteq V\cup E^H$ denotes the combined seed set of KHG entities and hyperedges, $\mathrm{dist}_{G_{\mathrm{BKG}}}(\cdot,\cdot)$ denotes the shortest-path distance on the bipartite knowledge graph $G_{\mathrm{BKG}}$, and $\rho$ denotes the expansion radius controlling how far the neighborhood grows.

\paragraph{Evidence Scoring and Fusion}
We organize knowledge evidence into separate entity and hyperedge rankings. Directly retrieved candidates take precedence: entities follow their retrieval order, and hyperedges follow the MMR selection sequence. Additional nodes from $\mathcal N_\rho(\mathcal E_0)$ extend their corresponding rankings in descending order of query relevance:
\begin{equation*}
  s_{\mathrm{text}}(t_q,u)
  =\cos\!\left(f_{\mathrm{text}}(t_q),f_{\mathrm{text}}(u)\right),
\end{equation*}
where $u$ denotes an entity or knowledge hyperedge in the expanded neighborhood.

Case-level evidence combines the RRF-ranked cases $R_P(t_q)$ with clinical references linked to the retrieved EEG windows. Cases in $R_P(t_q)$ retain their RRF order and precede the additional EEG-linked cases, which are ranked by their strongest window-level match to the query:
\begin{equation*}
  s_{\mathrm{EEG}}(p)
  =\max_{\substack{x\in R_{\mathrm{EEG}}(x_q)\\\operatorname{case}(x)=p}}
  s_r\!\left(f_{\mathrm{EEG}}(x_q),f_{\mathrm{EEG}}(x)\right),
\end{equation*}
where $\operatorname{case}(x)$ denotes the clinical case associated with EEG window $x$. The additional cases are ordered by decreasing $s_{\mathrm{EEG}}(p)$.

\paragraph{Context Construction}
We recover the source passages associated with the directly retrieved KHG entities and hyperedges:
\begin{equation*}
  \mathcal D_q
  =\bigcup_{u\in R_V(t_q)\cup R_H(t_q)}
  \operatorname{src}(u),
\end{equation*}
where $\operatorname{src}(u)\subseteq D$ denotes the source chunks associated with knowledge node $u$. The distinct passages are ranked by cosine similarity between their embeddings and the full query embedding.

We assemble the ranked evidence into $K^\ast$ under a total token budget $B$, with separate allocations for entities, facts, source passages, and clinical evidence. Within the clinical allocation, an EEG match and its case narrative receive initial priority, with ranked case narratives taking precedence over further EEG windows.

The context is organized by evidence type. Each reference case is presented through a shared clinical narrative and its associated EEG-window descriptions, connecting the case background with recording intervals, conditions, and event annotations. Source passages retain their provenance, providing the documentary context for the retrieved medical knowledge.

\subsection{Answer Generation}

For questions accompanied by an EEG recording, the waveform guides retrieval, while the language model receives the clinical context of the matched reference windows in textual form. Let $t_q$ denote the question and any accompanying clinical text, and let $p_{\mathrm{gen}}$ denote the generation instruction for the role specified in the query. Given the retrieved evidence $K^\ast$, the language model $\pi$ generates an answer $y^\ast$ as
\begin{equation*}
  y^\ast\sim \pi\!\left(y\mid t_q,K^\ast,p_{\mathrm{gen}}\right).
\end{equation*}
For case-specific questions, the model is instructed to ground claims about the patient in the observations supplied with the question. Retrieved medical knowledge and findings from reference cases provide context for interpreting those observations.

\section{EEG Clinical QA Benchmark}
\label{sec:benchmark}

\subsection{Data Sources}

The benchmark draws on seven public EEG datasets from OpenNeuro, covering epilepsy, depression, Parkinson's disease, Alzheimer's disease and frontotemporal dementia, sleep deprivation, first-episode psychosis, and mild traumatic brain injury~\cite{ds005873:1.1.0,ds003478:1.1.0,ds003509:1.1.0,ds004504:1.0.8,ds004902:1.0.8,ds003944:1.0.1,ds005114:1.0.0}. Each retained recording constitutes one case. We linked the EEG to the participant information, clinical assessments, and recording conditions available in the source study. This preserves distinctions between recordings from the same participant, such as whether a participant with Parkinson's disease was on or off medication at the time of recording.

We resampled the EEG to 128 Hz and segmented it into nonoverlapping 30-second windows, retaining the channel configuration defined for each domain. Each window retains its position in the original recording, so that events annotated by the source study can be linked to the intervals in which they occurred. In the epilepsy dataset, for example, cohort membership describes the participant, whereas the seizure annotations establish whether a particular window overlaps a recorded seizure.

A separate corpus of 14 documents provides the medical knowledge used in the benchmark. It comprises clinical guidelines, EEG recording and interpretation standards, and research articles relevant to the study domains~\cite{hall2021adhd,ACS2024TBI,VADoD2022MDD,kapur2017clinical,peltola2023joint,NICE2025epilepsy,APA2021schizophrenia,Sinha2016EEG,CDC2003MTBI,deBie2025PD,ICSD3TR2023insomnia,atri2025detecd,miltiadous2023eeg,xiang2024sleep}. These documents supply the basis for general knowledge questions and for questions that ask how a clinical principle applies to the observations in a case.

\subsection{Question-Answer Pair Construction}

We constructed each question family around a fact that could be verified in the source materials. For questions about a particular case, the reference answer was established from clinical observations, recording conditions, or annotated EEG events. For questions about clinical knowledge, we identified the passage in the literature that supported the answer. Questions involving the application of knowledge drew on both sources: the case supplied the relevant clinical circumstances, and the literature provided the basis for interpreting them.

With the target fact established, we used GPT-6 Astra to draft questions and answers from different role perspectives. For case-based questions, we selected the observations to disclose according to what the question asked the respondent to infer. The supporting source was retained for verification, and each draft was checked against it before expert review.

\subsection{Clinical Domains, Roles, and Tasks}

Across the seven study domains, we use a paired design to compare how well a system answers questions posed from different perspectives. Each question family addresses the same underlying fact from five perspectives: physician, nurse, clinical intern, patient, and researcher. The wording and emphasis vary with the role, while the clinical information and any EEG segment supplied with the question remain the same within a family. Each role is therefore represented across the same set of facts and cases.

Each domain contributes 20 question families with the same mix of four question types: four on foundational knowledge, eight on case reasoning, four linked to an EEG segment, and four on applying clinical knowledge to a case. This gives each domain equal weight and keeps the mix of question types comparable across domains. We also characterize questions by their clinical purpose. The 15 tasks represented in the benchmark include monitoring and documentation, diagnostic inference, clinical interpretation, and medication safety. Whereas question type describes the evidence a question draws on, clinical task describes what the respondent is being asked to do with that evidence.

\subsection{Expert Review and Quality Control}

Clinical experts reviewed the questions and reference answers for clinical relevance and correctness, considering the information provided in each question and the conditions under which the answer applies. Each expert assessed a distinct subset of items and assigned an acceptance decision to each. The final benchmark comprises 700 accepted questions.

\subsection{Data Splitting and Leakage Prevention}

We partitioned the clinical data by participant. The test set comprises the 87 participants whose recordings are associated with at least one benchmark question; the remaining 489 participants form the reference population for case and EEG retrieval. The partition applies to all visits and recordings from each participant. Questions about general EEG knowledge are not associated with a participant. We used publications accompanying the source datasets for their study methods and cohort-level findings. In the publication describing the dementia cohort~\cite{miltiadous2023eeg}, we excluded the participant-level table from the retrieval corpus to preserve the separation between individual case records and shared medical knowledge. Additional partition statistics and checks for duplicate recordings are provided in Appendix~\ref{app:split-audit}.

\subsection{Benchmark Statistics}

The seven source datasets comprise 2,353 recordings from 576 participants, yielding 694,386 EEG windows after preprocessing. Of these, 2,003 recordings and 564,011 windows form the reference collection. Although recording volume varies across domains, each domain contributes 100 benchmark questions.

Table~\ref{tab:benchmark-composition} summarizes the 700 questions by type. The 560 questions tied to individual cases draw on 97 recordings from 87 participants. The EEG-linked subset uses 28 distinct query windows, four per domain, each shared by the five role variants in its question family.

\begin{table}[t]
  \caption{Composition of the EEG clinical QA benchmark by question type. Some cases and participants appear in more than one question type.}
  \label{tab:benchmark-composition}
  \centering
  \small
  \setlength{\tabcolsep}{7pt}
  \begin{tabular}{@{}lccccc@{}}
    \toprule
    \textbf{Question type} & \textbf{Questions} & \textbf{Families} & \textbf{Cases} & \textbf{Participants} & \makecell{\textbf{Query EEG}\\\textbf{windows}} \\
    \midrule
    Foundational knowledge & 140 & 28 & --- & --- & --- \\
    Case reasoning         & 280 & 56 & 55  & 54  & --- \\
    EEG-linked             & 140 & 28 & 26  & 26  & 28  \\
    Knowledge application  & 140 & 28 & 27  & 27  & --- \\
    \midrule
    \textbf{Total}         & 700 & 140 & 97 & 87  & 28  \\
    \bottomrule
  \end{tabular}
\end{table}

\section{Experimental Setup}
\label{sec:setup}

Our experiments investigate how hierarchical retrieval supports EEG clinical question answering through five research questions:
\textbf{RQ1}: How does EEG-MedRAG compare with existing methods in answer quality across clinical domains and language models?
\textbf{RQ2}: How effectively does the framework retrieve relevant evidence and cover the knowledge needed to answer a query?
\textbf{RQ3}: How do the framework's components contribute to performance?
\textbf{RQ4}: How does performance vary across question types and clinical roles?
\textbf{RQ5}: What strengths and error patterns emerge across clinical QA tasks?
\subsection{Evaluation Protocol}
\label{sec:eval-protocol}

All methods are evaluated on the same benchmark questions and reference answers. Each query contains the question, its clinical role, and the available masked case information; EEG-linked queries additionally specify a waveform for retrieval. The reference case and EEG repositories exclude all records from test subjects, while the knowledge corpus is shared across methods. Comparisons within each answer-model group use a common generation prompt and evidence budget. For each method, the retrieved context is reused across answer models, allowing retrieval quality and answer quality to be assessed separately.

We report question-level averages overall and by clinical domain, question type, role, and task. Uncertainty is estimated using 95\% confidence intervals from 2,000 stratified cluster-bootstrap samples. Case-linked questions are clustered by subject, and questions without a case are clustered by question family, with strata defined by domain and case linkage. Method comparisons use paired scores on the same questions. For EEG retrieval, repeated role variants are grouped by query window, and uncertainty is estimated by resampling subjects within domains.

\subsection{Baselines}

We compare EEG-MedRAG with four baselines: NaiveGeneration, StandardRAG, HyperGraphRAG, and TimeRAG. All methods follow the evaluation protocol described in Section~\ref{sec:eval-protocol}.

\textbf{NaiveGeneration}~\cite{achiam2023gpt} generates answers from the question, role instruction, and available case information, relying on the language model's internal knowledge without external retrieval.

\textbf{StandardRAG}~\cite{gao2023retrieval} retrieves text passages from the medical literature corpus using dense embeddings. The question and accompanying case information form the query, and passages ranked by cosine similarity provide the evidence for answer generation.

\textbf{HyperGraphRAG}~\cite{luo2025hypergraphrag} combines entity and hyperedge retrieval over the medical knowledge hypergraph shared with EEG-MedRAG. It assembles retrieved entities, relational statements, and associated source passages into the generation context.

\textbf{TimeRAG}~\cite{yang2024timerag} uses representative time-series retrieval to provide EEG evidence for question answering. Reference windows are clustered within each clinical domain, with the window nearest each cluster center serving as its representative. Multivariate DTW ranks these representatives against the query EEG. The retrieved windows contribute their associated case narratives, recording conditions, and event annotations to answer generation. Questions without an EEG input are answered using the shared generator without retrieved evidence.

\subsection{Evaluation Metrics}

We evaluate answer quality and evidence retrieval at complementary levels. Let $\mathcal Q$ denote the benchmark questions, $y_q$ the answer to query $q$, $\mathcal G_q$ its set of acceptable reference answers, and $K_q^\ast$ the retrieved context used for generation.

\paragraph{Answer quality}
Answer correctness (AC) measures whether a response correctly and completely addresses the question. A fixed LLM judge assigns a binary score $J_{\mathrm{AC}}(q,y_q,\mathcal G_q)$ based on the question, role, available case information, and reference answers, without method or model identities, retrieved context, or intermediate reasoning. The judgment accepts equivalent formulations while accounting for numerical conditions, negation, and material factual errors. We report
\begin{equation*}
  \mathrm{AC}=\frac{1}{|\mathcal Q|}\sum_{q\in\mathcal Q}
  J_{\mathrm{AC}}(q,y_q,\mathcal G_q).
\end{equation*}
Token-level F1 complements AC by measuring lexical agreement with the reference answers. It is computed from multiset token overlap, taking the highest score over acceptable references and then averaging across questions. The token normalization and scoring rules are given in Appendix~\ref{app:eval-details}.

\paragraph{Evidence coverage}
Evidence coverage (EC) measures how much of the knowledge needed to answer a query is supported by the retrieved context. Following the claim-level evaluation approach of RAGChecker~\cite{ru2024ragchecker}, we construct a set $\mathcal T_q$ of atomic knowledge requirements from the question and its acceptable reference answers. These requirements capture external knowledge beyond the facts supplied in the query. They are extracted and checked by the LLM judge before method comparison, then held fixed across methods. For a question with a nonempty requirement set,
\begin{equation*}
  \mathrm{EC}_q=\frac{1}{|\mathcal T_q|}
  \sum_{t\in\mathcal T_q}J_{\mathrm{sup}}(t,K_q^\ast),
\end{equation*}
where $J_{\mathrm{sup}}$ indicates whether the context supports requirement $t$. We average EC over eligible foundational-knowledge and knowledge-application questions.

\paragraph{Context relevance}
Context relevance (CRel) estimates the proportion of retrieved content that helps answer the query. Following the context-relevance approach of RAGAs~\cite{es2024ragas}, we divide the retrieved context into sentence- or line-level units. For each nonempty context, we draw a uniform sample $\mathcal S_q$ of up to 24 units without replacement. The judge assesses the relevance of each sampled unit in light of the full context, yielding
\begin{equation*}
  \mathrm{CRel}_q=\frac{1}{|\mathcal S_q|}
  \sum_{u\in\mathcal S_q}J_{\mathrm{rel}}(q,u,K_q^\ast).
\end{equation*}
The judgment covers medical knowledge, reference-case information, and EEG-neighbor descriptions, using the query and its available case information. CRel is averaged across questions and reported alongside context length. Together, EC and CRel characterize the coverage and focus of the evidence supplied for answer generation.

\paragraph{EEG label agreement}
We assess the disease- and state-related information captured by waveform retrieval using label agreement at five (LA@5). Let $\ell(x)$ denote the recorded task label of EEG window $x$. For a query window $x_q$ with a known label,
\begin{equation*}
  \mathrm{LA@5}(x_q)=\frac{1}{5}
  \sum_{x\in R_{\mathrm{EEG}}(x_q)}
  \mathbb I[\ell(x)=\ell(x_q)],
\end{equation*}
where $R_{\mathrm{EEG}}(x_q)$ contains up to five neighbors before context selection. Missing neighbors and neighbors with unknown labels contribute no matches. We average over distinct labeled query windows and compare with the expected agreement under uniform sampling from the same-domain reference repository. The domain-specific labels and their availability are described in Appendix~\ref{app:eval-details}.

\subsection{Implementation Details}

We use GPT-4o-mini for knowledge extraction and query processing, and text-embedding-3-small for text representations. Source documents are divided into chunks of 1,200 tokens with an overlap of 100 tokens. Answer generation is evaluated with GPT-4o-mini, Gemini-2.5-Flash, and DeepSeek-V4-Flash. GPT-4.1-mini serves as the fixed judge for AC, EC, and CRel and constructs the shared knowledge requirements used in EC.

All methods use the same answer-generation template, with a decoding temperature of zero and a maximum of 4,096 output tokens. Retrieval-based methods operate under a common context budget of 32,000 tokens. For EEG-MedRAG, this budget is allocated to entities, facts, source passages, and clinical evidence, with 4,000, 6,000, 10,000, and 12,000 tokens, respectively.

For knowledge retrieval, we set $k_V=k_H=60$ and $k_{\mathrm{cand}}=240$, with an MMR trade-off of $\lambda=0.5$. Each case retains up to $k_{\mathrm{soft}}=8$ soft links to KHG nodes with cosine similarity above 0.2, and we set $\rho=1$ for knowledge expansion. Both case retrieval routes and the merged case list use a cutoff of $k_P=5$, with $c_{\mathrm{RRF}}=60$.

Each EEG-based benchmark question uses a single 30-second query window at 128 Hz. We compute $n=1920$ PAA values per channel, with each value averaging two consecutive samples ($15.625\,\mathrm{ms}$). We retrieve the $K=5$ nearest reference EEG windows within the same clinical domain as the query using multivariate DTW with a shared temporal alignment path and a channel-summed L1 local cost. The temporal tolerance is set to $\Delta=0.25\,\mathrm{s}$, corresponding to a Sakoe--Chiba radius of $r=16$ at this resolution.

All retrieval baselines operate on the same text chunks, embeddings, knowledge hypergraph, or PAA windows under the shared 32,000-token context budget. StandardRAG uses a cosine similarity threshold of $0.5$ for retrieval. HyperGraphRAG uses its default hybrid retrieval configuration ($k=60$, cosine threshold $0.2$, and 4,000-token per-path sub-budgets). TimeRAG partitions each domain's $n=1920$ PAA windows into $\lceil\sqrt{N}\rceil$ clusters via mini-batch $k$-means and retrieves the $K=5$ nearest cluster representatives using unconstrained path-normalized Euclidean DTW.

\section{Experimental Results}
\label{sec:results}

\subsection{Overall Performance}
\label{sec:overall-results}

Table~\ref{tab:performance-by-domain} reports answer correctness (AC) and token-level F1 across the seven clinical domains and three generation backbones, addressing \textbf{RQ1}.

\begin{table}[!hbp]
  \caption{Answer correctness (AC) and token-level F1 across seven clinical domains, reported as percentages. Overall scores are averages over all 700 questions. For each generation model, the best score in each column is shown in bold. AD = Alzheimer's disease; FTD = frontotemporal dementia.}
  \label{tab:performance-by-domain}
  \centering
  \footnotesize
  \setlength{\tabcolsep}{2.6pt}
  \begin{tabular}{@{}lcc|cc|cc|cc|cc|cc|cc|cc@{}}
    \toprule
    \multirow{2}{*}{\textbf{Method}} &
    \multicolumn{2}{c}{\textbf{Epilepsy}} &
    \multicolumn{2}{c}{\makecell{\textbf{Parkinson's}\\\textbf{disease}}} &
    \multicolumn{2}{c}{\textbf{AD/FTD}} &
    \multicolumn{2}{c}{\textbf{Depression}} &
    \multicolumn{2}{c}{\makecell{\textbf{Sleep}\\\textbf{deprivation}}} &
    \multicolumn{2}{c}{\textbf{Mild TBI}} &
    \multicolumn{2}{c}{\makecell{\textbf{First-episode}\\\textbf{psychosis}}} &
    \multicolumn{2}{c}{\textbf{Overall}} \\
    \cmidrule(lr){2-3}\cmidrule(lr){4-5}\cmidrule(lr){6-7}\cmidrule(lr){8-9}
    \cmidrule(lr){10-11}\cmidrule(lr){12-13}\cmidrule(lr){14-15}\cmidrule(l){16-17}
    & \textbf{AC} & \textbf{F1}
    & \textbf{AC} & \textbf{F1}
    & \textbf{AC} & \textbf{F1}
    & \textbf{AC} & \textbf{F1}
    & \textbf{AC} & \textbf{F1}
    & \textbf{AC} & \textbf{F1}
    & \textbf{AC} & \textbf{F1}
    & \textbf{AC} & \textbf{F1} \\
    \midrule
    \multicolumn{17}{c}{\textbf{\textit{GPT-4o-mini}}} \\
    \midrule
    NaiveGeneration & 70.00 & 47.13 & 59.00 & 39.68 & 73.00 & 44.15 & 78.00 & 42.34 & 67.00 & 41.62 & 71.00 & 38.42 & 77.00 & 39.31 & 70.71 & 41.81 \\
    StandardRAG & 70.00 & 50.43 & 77.00 & 46.26 & 78.00 & 48.70 & 80.00 & 45.84 & 78.00 & 47.50 & 77.00 & 41.25 & 84.00 & 40.89 & 77.71 & 45.84 \\
    HyperGraphRAG & 73.00 & 53.54 & 81.00 & 46.58 & 81.00 & 48.12 & 77.00 & 45.44 & 77.00 & 47.25 & 76.00 & 42.09 & 82.00 & 41.26 & 78.14 & 46.32 \\
    TimeRAG & 76.00 & 51.74 & 55.00 & 39.51 & 73.00 & 45.46 & 79.00 & 41.56 & 65.00 & 43.44 & 70.00 & 39.49 & 77.00 & 38.20 & 70.71 & 42.77 \\
    \textbf{EEG-MedRAG (ours)} & \textbf{81.00} & \textbf{54.31} & \textbf{86.00} & \textbf{47.56} & \textbf{83.00} & \textbf{49.77} & \textbf{83.00} & \textbf{46.44} & \textbf{79.00} & \textbf{49.06} & \textbf{85.00} & \textbf{43.39} & \textbf{89.00} & \textbf{43.92} & \textbf{83.71} & \textbf{47.78} \\
    \midrule
    \multicolumn{17}{c}{\textbf{\textit{Gemini-2.5-Flash}}} \\
    \midrule
    NaiveGeneration & 70.00 & 19.72 & 70.00 & 23.96 & 78.00 & 23.33 & 75.00 & 24.40 & 76.00 & 23.77 & 79.00 & 20.33 & 82.00 & 16.92 & 75.71 & 21.78 \\
    StandardRAG & 81.00 & 21.42 & 81.00 & 24.79 & 82.00 & 24.92 & 80.00 & 27.15 & 80.00 & 27.19 & 80.00 & 19.87 & 85.00 & 18.56 & 81.29 & 23.41 \\
    HyperGraphRAG & 82.00 & 19.77 & 84.00 & 41.03 & 86.00 & 27.20 & 81.00 & 28.44 & 80.00 & 40.81 & 83.00 & 30.43 & 90.00 & 31.52 & 83.71 & 31.32 \\
    TimeRAG & 75.00 & 22.05 & 68.00 & 23.40 & 77.00 & 26.65 & 72.00 & 26.04 & 78.00 & 30.11 & 80.00 & 22.64 & 77.00 & 23.49 & 75.29 & 24.91 \\
    \textbf{EEG-MedRAG (ours)} & \textbf{86.00} & \textbf{44.68} & \textbf{85.00} & \textbf{48.16} & \textbf{87.00} & \textbf{40.39} & \textbf{87.00} & \textbf{42.03} & \textbf{81.00} & \textbf{42.85} & \textbf{84.00} & \textbf{38.28} & \textbf{93.00} & \textbf{46.57} & \textbf{86.14} & \textbf{43.28} \\
    \midrule
    \multicolumn{17}{c}{\textbf{\textit{DeepSeek-V4-Flash}}} \\
    \midrule
    NaiveGeneration & 63.00 & 38.10 & 64.00 & 41.36 & 73.00 & 51.16 & 78.00 & 50.92 & 78.00 & 38.55 & 73.00 & 32.16 & 75.00 & 22.66 & 72.00 & 39.27 \\
    StandardRAG & 72.00 & 38.22 & 77.00 & 38.08 & 79.00 & 58.66 & 68.00 & 39.60 & 78.00 & 37.18 & \textbf{81.00} & 28.10 & 72.00 & 27.37 & 75.29 & 38.17 \\
    HyperGraphRAG & 53.00 & 25.32 & 69.00 & 31.27 & 80.00 & 33.13 & 33.00 & 17.55 & 73.00 & 36.23 & 78.00 & 31.94 & 73.00 & 23.94 & 65.57 & 28.48 \\
    TimeRAG & 72.00 & 37.41 & 73.00 & 38.84 & 72.00 & 33.60 & 72.00 & 36.28 & 78.00 & 39.34 & 78.00 & 33.51 & 77.00 & 29.06 & 74.57 & 35.43 \\
    \textbf{EEG-MedRAG (ours)} & \textbf{81.00} & \textbf{51.99} & \textbf{86.00} & \textbf{49.77} & \textbf{81.00} & \textbf{59.71} & \textbf{81.00} & \textbf{53.26} & \textbf{80.00} & \textbf{66.38} & \textbf{81.00} & \textbf{45.54} & \textbf{79.00} & \textbf{51.68} & \textbf{81.29} & \textbf{54.05} \\
    \bottomrule
  \end{tabular}
\end{table}

Across all 700 benchmark questions, EEG-MedRAG consistently establishes the strongest overall performance regardless of the underlying language model. With GPT-4o-mini, it reaches 83.71\% AC and 47.78 F1, surpassing the strongest baseline, HyperGraphRAG, by +5.57 percentage points in AC (95\% paired bootstrap CI [+2.78, +8.54]) and outperforming NaiveGeneration and TimeRAG by +13.00 points in AC ([+8.88, +17.09] and [+9.44, +16.49]). With Gemini-2.5-Flash, EEG-MedRAG attains the highest accuracy in the entire evaluation at 86.14\% AC while lifting F1 to 43.28---a gain of +11.96 to +21.50 points over all baselines, with every paired F1 confidence interval strictly positive. The widest overall margins emerge on DeepSeek-V4-Flash, where EEG-MedRAG reaches 81.29\% AC and 54.05 F1, exceeding every baseline by +6.00 to +15.71 percentage points in AC and +14.77 to +25.57 points in F1 (all paired 95\% CIs strictly above zero; complete paired intervals appear in Appendix~\ref{app:additional-results}).

Comparing the baseline families clarifies why hierarchical multimodal retrieval is necessary for EEG clinical QA. Relying solely on parametric knowledge (NaiveGeneration) or on waveform-only retrieval (TimeRAG) leaves substantial accuracy gaps across all three backbones. Although TimeRAG improves over NaiveGeneration in epilepsy---where seizure-related queries directly benefit from reference waveforms---it lacks external medical knowledge and case context for non-EEG questions, keeping its overall AC near NaiveGeneration. Adding literature retrieval (StandardRAG) yields clear gains on GPT-4o-mini and Gemini-2.5-Flash, and organizing that literature into $n$-ary facts (HyperGraphRAG) further improves overall accuracy on both backbones. Yet knowledge-only retrieval still plateaus short of EEG-MedRAG because clinical questions often hinge on combining general medical principles with comparable patient cases and physiological evidence. The cross-backbone patterns reinforce this advantage in two ways. On Gemini-2.5-Flash, where baselines achieve respectable AC but low lexical alignment (21.78--31.32 F1) due to verbose, loosely grounded responses, EEG-MedRAG's structured evidence context sharpens precision and raises F1 by nearly 12 points over HyperGraphRAG. On DeepSeek-V4-Flash, long uncurated graph contexts in HyperGraphRAG trigger provider content filters on 149 queries (concentrated in depression and epilepsy), depressing its overall AC to 65.57\%, whereas EEG-MedRAG's budgeted four-section context markedly reduces filtering (40 queries) and sustains both high correctness (81.29\% overall; 86.21\% on unfiltered queries) and concise clinical phrasing (54.05 F1).

The performance gains generalize broadly across clinical conditions rather than concentrating in a few favorable cohorts. EEG-MedRAG ranks first in F1 across all 21 domain--model combinations and achieves the highest AC in 20 of 21 settings, tying StandardRAG in the remaining one (mild TBI on DeepSeek-V4-Flash, 81.00\%). Notably, EEG-MedRAG excels not only in domains with rich diagnostic guidelines and structured cognitive scales---such as Parkinson's disease (86.00\% AC on both GPT-4o-mini and DeepSeek-V4-Flash) and AD/FTD (87.00\% AC on Gemini-2.5-Flash)---but also in clinically heterogeneous conditions where text-only baselines struggle, including mild TBI (+8.00 points AC over the best baseline on GPT-4o-mini) and first-episode psychosis (89.00\% AC on GPT-4o-mini and 93.00\% on Gemini-2.5-Flash). In epilepsy, EEG-MedRAG outperforms both TimeRAG and HyperGraphRAG across all three backbones (81.00\%, 86.00\%, and 81.00\% AC), showing that waveform matching and structured clinical knowledge reinforce each other rather than competing. To examine the retrieval mechanisms underlying these answer-level gains, we next evaluate evidence coverage and context relevance directly.

\subsection{Retrieval Effectiveness}

\begin{table}[t]
  \caption{Retrieval effectiveness by method. Each method's retrieved context is reused across generation models. Evidence coverage (EC) is averaged over 280 foundational knowledge and knowledge application questions. Context relevance (CRel), mean context length, and empty-context counts cover all 700 questions. Brackets give 95\% stratified cluster-bootstrap confidence intervals.}
  \label{tab:retrieval-quality}
  \centering
  \small
  \begin{tabular}{@{}lcccc@{}}
    \toprule
    \textbf{Method} & \textbf{EC (\%) [95\% CI]} & \textbf{CRel (\%) [95\% CI]} & \textbf{Mean tokens} & \textbf{Empty / 700} \\
    \midrule
    StandardRAG & 78.75 [71.07, 85.89] & 14.48 [12.46, 16.33] & 3,799 & 225 \\
    HyperGraphRAG & 94.73 [90.48, 98.10] & 13.52 [12.43, 14.51] & 17,378 & 81 \\
    TimeRAG & 0.00 [0.00, 0.00] & 9.57 [6.15, 12.82] & 948 & 560 \\
    \textbf{EEG-MedRAG (ours)} & \textbf{97.41} [93.81, 100.00] & \textbf{19.74} [18.56, 20.84] & 23,489 & 0 \\
    \bottomrule
  \end{tabular}
\end{table}

Table~\ref{tab:retrieval-quality} evaluates retrieval quality directly on the shared contexts supplied to the generation models, addressing \textbf{RQ2}. EEG-MedRAG achieves both the highest evidence coverage (97.41\% EC) and the highest context relevance (19.74\% CRel), with zero empty contexts across all 700 queries. Paired cluster-bootstrap comparisons confirm that these gains are statistically reliable against every baseline: EEG-MedRAG exceeds StandardRAG by +18.66 percentage points in EC (95\% CI [+12.14, +25.71]) and +5.27 points in CRel ([+3.78, +6.76]), and outperforms HyperGraphRAG by +2.68 points in EC ([+0.29, +5.50]) and +6.23 points in CRel ([+5.23, +7.14]). Notably, although EEG-MedRAG assembles a richer average context than HyperGraphRAG (23,489 versus 17,378 tokens), it avoids the typical coverage--relevance trade-off: budgeted four-section selection, MMR fact diversification, and cross-layer case grounding simultaneously expand knowledge coverage and raise the proportion of task-relevant evidence.

Disaggregating retrieval behavior by question type clarifies how these gains arise (full domain and question-type breakdowns appear in Appendix~\ref{app:additional-results}). StandardRAG's flat chunk retrieval misses nearly a fifth of required external knowledge (78.75\% EC) and returns empty contexts on 192 of 280 case-reasoning queries when query narratives fall below its similarity cutoff. HyperGraphRAG recovers most foundational facts (99.29\% EC on foundational knowledge) but still leaves 80 case-reasoning queries empty and drops to 12.53\% CRel on EEG-linked queries because it retrieves only general literature. Conversely, TimeRAG yields highly focused reference contexts on the 140 EEG-linked questions (47.84\% CRel at 4,741 mean tokens), confirming the relevance of waveform-matched patient cases, yet returns empty contexts on all 560 non-EEG queries (0.00\% EC). EEG-MedRAG bridges both regimes: it reaches 100.00\% EC on foundational knowledge, raises knowledge-application coverage from 90.18\% (HyperGraphRAG) to 94.82\% while lifting CRel from 19.29\% to 27.08\% with a shorter mean context (21,564 versus 24,577 tokens), and achieves the highest CRel in every clinical domain. This combination of near-complete knowledge coverage and focused multimodal context directly accounts for the consistent answer-level gains observed in Section~\ref{sec:overall-results}.

\subsection{Ablation Studies}
\label{sec:ablation}

\begin{table}[t]
  \caption{Leave-one-out ablation of EEG-MedRAG retrieval layers with GPT-4o-mini, reported as percentages across question types and all 700 queries. $\Delta$AC denotes the paired difference relative to the full framework (Variant $-$ Full) with 95\% stratified cluster-bootstrap confidence intervals. Best scores per column are shown in bold.}
  \label{tab:ablation-loo}
  \centering
  \footnotesize
  \setlength{\tabcolsep}{2.6pt}
  \begin{tabular}{@{}lcc|cc|cc|cc|cc|c@{}}
    \toprule
    \multirow{2}{*}{\textbf{Configuration}} &
    \multicolumn{2}{c}{\makecell{\textbf{Foundational}\\\textbf{knowledge (140)}}} &
    \multicolumn{2}{c}{\makecell{\textbf{Knowledge}\\\textbf{application (140)}}} &
    \multicolumn{2}{c}{\makecell{\textbf{Case}\\\textbf{reasoning (280)}}} &
    \multicolumn{2}{c}{\makecell{\textbf{EEG-}\\\textbf{linked (140)}}} &
    \multicolumn{2}{c}{\textbf{Overall (700)}} &
    \multirow{2}{*}{\makecell{\textbf{$\Delta$AC (pp)}\\\textbf{[95\% CI]}}} \\
    \cmidrule(lr){2-3}\cmidrule(lr){4-5}\cmidrule(lr){6-7}\cmidrule(lr){8-9}\cmidrule(lr){10-11}
    & \textbf{AC} & \textbf{F1}
    & \textbf{AC} & \textbf{F1}
    & \textbf{AC} & \textbf{F1}
    & \textbf{AC} & \textbf{F1}
    & \textbf{AC} & \textbf{F1} & \\
    \midrule
    \textbf{EEG-MedRAG (Full)} & \textbf{99.29} & 59.43 & \textbf{96.43} & \textbf{34.84} & 96.79 & 62.18 & \textbf{29.29} & 20.25 & \textbf{83.71} & \textbf{47.78} & --- \\
    w/o Knowledge (KHG) & 78.57 & 41.06 & 75.71 & 26.25 & \textbf{98.21} & \textbf{65.96} & 20.00 & \textbf{21.17} & 74.14 & 44.08 & $-9.57$ [$-13.36$, $-5.88$] \\
    w/o Case (CHG) & \textbf{99.29} & 59.49 & 91.43 & 33.30 & 93.57 & 57.81 & 13.57 & 16.56 & 78.29 & 44.99 & $-5.43$ [$-8.91$, $-2.36$] \\
    w/o EEG (EVD) & \textbf{99.29} & \textbf{59.86} & 91.43 & 34.81 & \textbf{98.21} & 61.02 & 12.14 & 16.00 & 79.86 & 46.54 & $-3.86$ [$-7.23$, $-0.83$] \\
    \bottomrule
  \end{tabular}
\end{table}

\begin{table}[t]
  \caption{Full EEG-MedRAG and its single-layer variants across three generation models. AC and F1 cover all 700 questions. Retrieval metrics are shared across models, with EC averaged over 280 knowledge questions and CRel and context length over all 700. For DeepSeek-V4-Flash, $^\dagger$ indicates that filtered responses scored zero (40 for Full and 50 for Knowledge-only).}
  \label{tab:ablation-single}
  \centering
  \small
  \setlength{\tabcolsep}{4pt}
  \begin{tabular}{@{}lcc|cc|cc|ccc@{}}
    \toprule
    \multirow{2}{*}{\textbf{Configuration}} &
    \multicolumn{2}{c}{\textbf{GPT-4o-mini}} &
    \multicolumn{2}{c}{\textbf{Gemini-2.5-Flash}} &
    \multicolumn{2}{c}{\textbf{DeepSeek-V4-Flash}} &
    \multicolumn{3}{c}{\textbf{Shared retrieval}} \\
    \cmidrule(lr){2-3}\cmidrule(lr){4-5}\cmidrule(lr){6-7}\cmidrule(l){8-10}
    & \textbf{AC} & \textbf{F1}
    & \textbf{AC} & \textbf{F1}
    & \textbf{AC} & \textbf{F1}
    & \textbf{EC (\%)} & \textbf{CRel (\%)} & \textbf{Tokens} \\
    \midrule
    \textbf{EEG-MedRAG (Full)} & \textbf{83.71} & \textbf{47.78} & \textbf{86.14} & \textbf{43.28} & \textbf{81.29}$^\dagger$ & \textbf{54.05} & \textbf{97.41} & \textbf{19.74} & 23,489 \\
    Knowledge-only (KHG) & 78.29 & 44.29 & 79.57 & 41.07 & 70.57$^\dagger$ & 37.08 & 95.45 & 13.05 & 29,764 \\
    Case-only (CHG) & 72.86 & 42.79 & 75.57 & 29.95 & 81.00 & 35.19 & 0.89 & 19.17 & 2,397 \\
    EEG-only (EVD) & 70.71 & 42.18 & 77.00 & 25.61 & 81.00 & 34.83 & 0.00 & 8.44 & 591 \\
    \bottomrule
  \end{tabular}
\end{table}

To answer \textbf{RQ3}, we evaluate component contributions from two complementary angles: disabling one retrieval layer at a time (Table~\ref{tab:ablation-loo}) and isolating each layer individually across all three generation backbones (Table~\ref{tab:ablation-single}). In the leave-one-out study with GPT-4o-mini, removing any single layer causes a statistically reliable drop in both overall AC and F1. Disabling the knowledge hypergraph (\emph{w/o Knowledge}) produces the steepest overall decline ($-9.57$ percentage points AC, 95\% CI [$-13.36$, $-5.88$]; $-3.70$ F1, [$-5.84$, $-1.61$]), lowering both foundational knowledge and knowledge application by more than 20 points while also reducing EEG-linked accuracy from 29.29\% to 20.00\%. Removing the EEG repository (\emph{w/o EEG}) leaves foundational knowledge untouched (99.29\%) yet cuts EEG-linked accuracy by more than half, from 29.29\% to 12.14\% ($-3.86$ points overall AC, [$-7.23$, $-0.83$]). Disabling the case hypergraph (\emph{w/o Case}) yields the second-largest overall drop ($-5.43$ points AC, [$-8.91$, $-2.36$]; $-2.78$ F1, [$-4.24$, $-1.40$]) and hurts three distinct question types simultaneously---case reasoning, knowledge application, and EEG-linked inference (dropping to 13.57\%)---confirming CHG's bridging role between clinical narratives, waveform matches, and structured domain knowledge.

The single-layer evaluation in Table~\ref{tab:ablation-single} demonstrates that these layers act synergistically rather than as interchangeable evidence sources. On GPT-4o-mini and Gemini-2.5-Flash, the full framework decisively outperforms every isolated layer by +5.43 to +13.00 percentage points in AC (all paired 95\% CIs strictly positive). This synergy is most striking on EEG-linked questions with GPT-4o-mini: none of the isolated layers exceeds 10.00\% AC (\emph{Knowledge-only} 8.57\%, \emph{Case-only} 10.00\%, \emph{EEG-only} 8.57\%), whereas combining all three nearly triples accuracy to 29.29\%. Retrieval metrics exhibit the same complementarity. Although \emph{Knowledge-only} consumes more context tokens than \emph{Full} (29,764 versus 23,489 tokens) because the unused clinical budget is redistributed into KHG, it still trails \emph{Full} in both EC (95.45\% versus 97.41\%) and CRel (13.05\% versus 19.74\%), showing that case-seeded soft links and clinical partitioning improve both coverage and relevance. Meanwhile, \emph{Case-only} and \emph{EEG-only} provide compact clinical references (0.89\% and 0.00\% EC) but cannot ground literature-dependent questions.

On DeepSeek-V4-Flash, \emph{Full} likewise leads all single-layer variants in both AC and F1, outperforming \emph{Knowledge-only} by +10.71 percentage points in AC (95\% CI [+5.54, +15.21]) and exceeding every isolated layer by +16.97 to +19.21 points in F1 (54.05 versus 34.83--37.08). Despite 40 provider-filtered queries on long knowledge-rich contexts, \emph{Full} still attains the highest overall AC (81.29\%, rising to 86.21\% on unfiltered queries). Taken together, both ablation designs confirm that KHG, CHG, and EVD play distinct, mutually reinforcing roles across the clinical QA spectrum.

\subsection{Question-Type and Role Analysis}
\label{sec:type-role}

Stratifying performance by question type and clinical role illuminates how EEG-MedRAG handles differing evidentiary demands and user perspectives, addressing \textbf{RQ4}. Table~\ref{tab:question-type-results} shows clear separation across the four benchmark question types.

\suppressfloats[t]
\begin{table}[htp]
  \caption{Answer correctness (AC) and token-level F1 (\%) stratified by benchmark question type across three generation models. Best scores per column within each model group are shown in bold.}
  \label{tab:question-type-results}
  \centering
  \small
  \setlength{\tabcolsep}{4.5pt}
  \begin{tabular}{@{}lcc|cc|cc|cc@{}}
    \toprule
    \multirow{2}{*}{\textbf{Method}} &
    \multicolumn{2}{c}{\makecell{\textbf{Foundational}\\\textbf{knowledge (140)}}} &
    \multicolumn{2}{c}{\makecell{\textbf{Knowledge}\\\textbf{application (140)}}} &
    \multicolumn{2}{c}{\makecell{\textbf{Case}\\\textbf{reasoning (280)}}} &
    \multicolumn{2}{c}{\makecell{\textbf{EEG-}\\\textbf{linked (140)}}} \\
    \cmidrule(lr){2-3}\cmidrule(lr){4-5}\cmidrule(lr){6-7}\cmidrule(l){8-9}
    & \textbf{AC} & \textbf{F1}
    & \textbf{AC} & \textbf{F1}
    & \textbf{AC} & \textbf{F1}
    & \textbf{AC} & \textbf{F1} \\
    \midrule
    \multicolumn{9}{c}{\textbf{\textit{GPT-4o-mini}}} \\
    \midrule
    NaiveGeneration & 77.14 & 40.86 & 79.29 & 27.41 & \textbf{97.14} & 62.99 & 2.86 & 14.78 \\
    StandardRAG & 95.00 & 56.63 & 89.29 & 34.82 & 95.71 & 62.04 & 12.86 & 13.67 \\
    HyperGraphRAG & 98.57 & \textbf{60.95} & 92.14 & 34.13 & 92.86 & 61.33 & 14.29 & 13.87 \\
    TimeRAG & 73.57 & 41.19 & 74.29 & 26.42 & 96.79 & \textbf{64.32} & 12.14 & 17.61 \\
    \textbf{EEG-MedRAG (ours)} & \textbf{99.29} & 59.43 & \textbf{96.43} & \textbf{34.84} & 96.79 & 62.18 & \textbf{29.29} & \textbf{20.25} \\
    \midrule
    \multicolumn{9}{c}{\textbf{\textit{Gemini-2.5-Flash}}} \\
    \midrule
    NaiveGeneration & 70.71 & 19.70 & 79.29 & 16.21 & \textbf{98.93} & 31.50 & 30.71 & 9.96 \\
    StandardRAG & 94.29 & 28.26 & 87.86 & 20.29 & 95.71 & 30.60 & 32.86 & 7.33 \\
    HyperGraphRAG & 98.57 & 42.62 & 91.43 & 28.75 & 95.00 & 34.70 & 38.57 & 15.81 \\
    TimeRAG & 75.00 & 23.14 & 72.14 & 17.00 & 98.57 & 31.95 & 32.14 & 20.51 \\
    \textbf{EEG-MedRAG (ours)} & \textbf{100.00} & \textbf{48.04} & \textbf{95.00} & \textbf{34.16} & 96.79 & \textbf{47.47} & \textbf{42.14} & \textbf{39.26} \\
    \midrule
    \multicolumn{9}{c}{\textbf{\textit{DeepSeek-V4-Flash}}} \\
    \midrule
    NaiveGeneration & 78.57 & 40.09 & 75.00 & 25.84 & \textbf{96.79} & 59.64 & 12.86 & 11.16 \\
    StandardRAG & 79.29 & 42.05 & 81.43 & 24.46 & 95.00 & 56.59 & 25.71 & 11.18 \\
    HyperGraphRAG & 66.43 & 35.27 & 61.43 & 14.29 & 85.71 & 41.22 & 28.57 & 10.42 \\
    TimeRAG & 80.71 & 32.81 & 77.86 & 21.30 & 96.07 & 51.15 & 22.14 & 20.76 \\
    \textbf{EEG-MedRAG (ours)} & \textbf{82.86} & \textbf{52.39} & \textbf{90.00} & \textbf{33.47} & 95.36 & \textbf{75.93} & \textbf{42.86} & \textbf{32.52} \\
    \bottomrule
  \end{tabular}
\end{table}

\begin{figure*}[t]
  \centering
  \includegraphics[width=\textwidth]{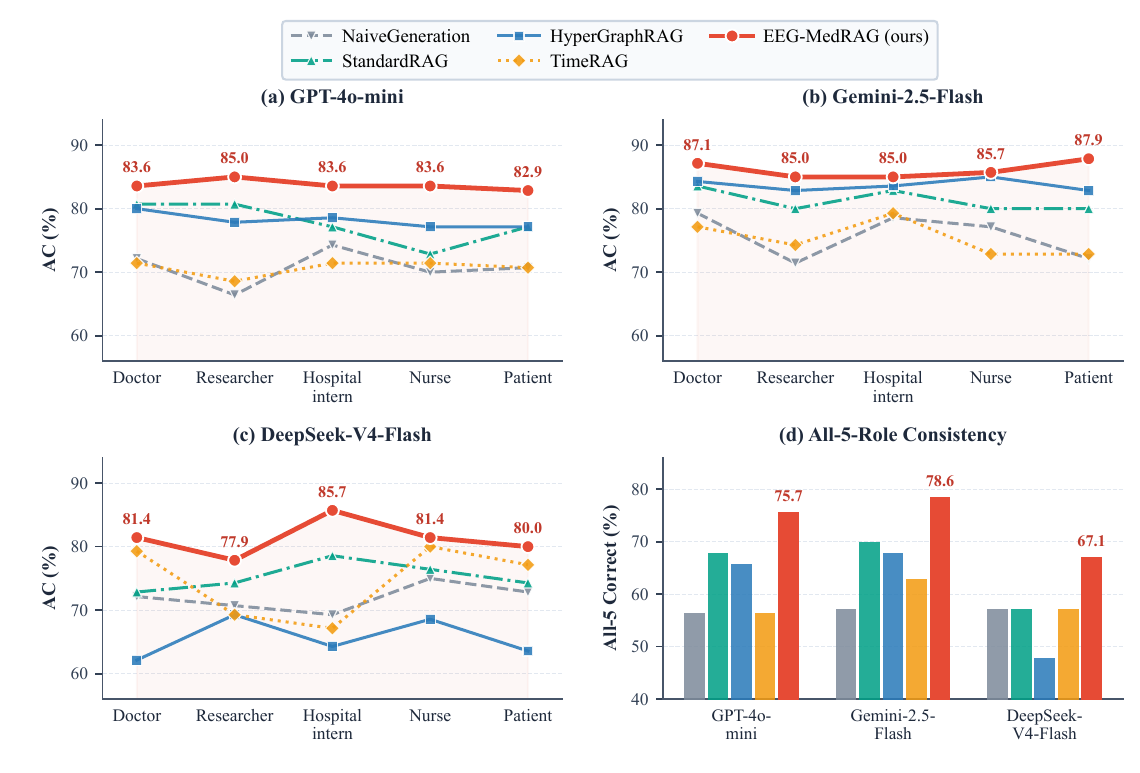}
  \caption{Cross-role performance and paired-family consistency across three generation backbones: (a)--(c) answer correctness (AC, \%) across five clinical roles under GPT-4o-mini, Gemini-2.5-Flash, and DeepSeek-V4-Flash; (d) percentage of the 140 paired question families in which all five role variants are answered correctly.}
  \Description{Four-panel figure showing answer correctness across five clinical roles as connected profile plots under GPT-4o-mini, Gemini-2.5-Flash, and DeepSeek-V4-Flash in panels (a) to (c), and all-five-role family consistency as a grouped bar chart across the three models in panel (d).}
  \label{fig:ac-by-role-and-model}
\end{figure*}

\begin{figure*}[t]
  \centering
  \includegraphics[width=\textwidth]{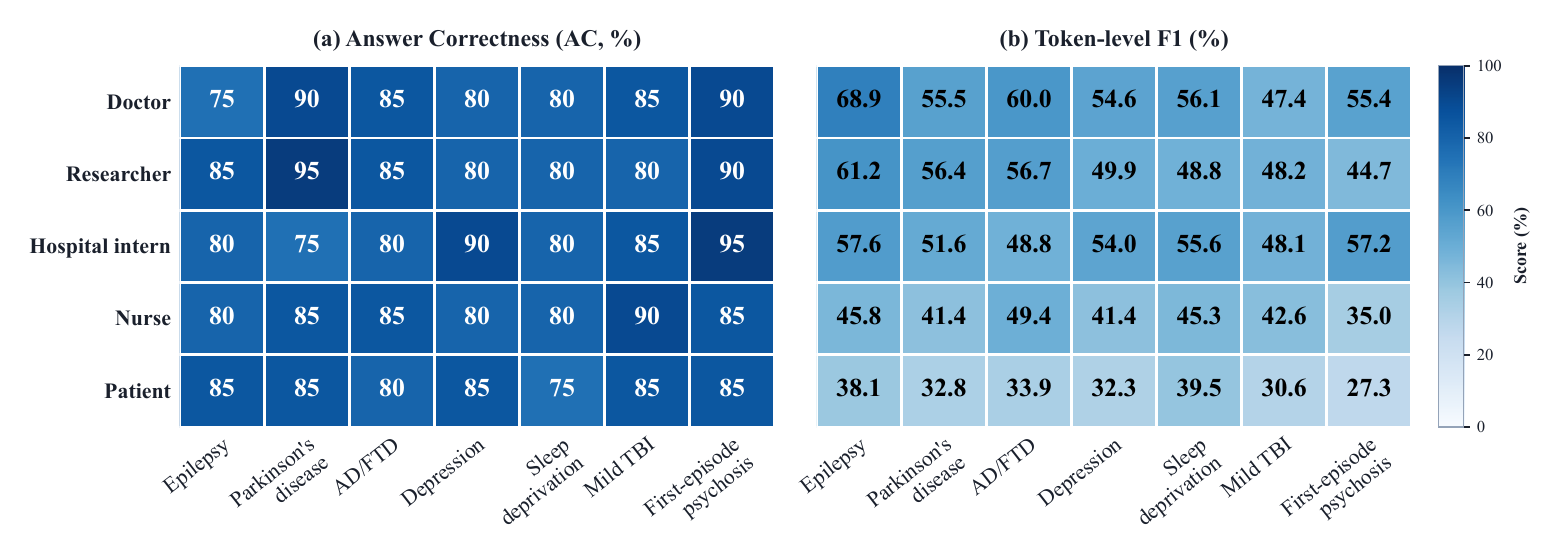}
  \caption{EEG-MedRAG performance with GPT-4o-mini by clinical role and domain. Panels (a) and (b) show answer correctness (AC) and token-level F1, respectively, as percentages. Questions asked from the patient's perspective yield the lowest F1 in each of the seven domains.}
  \Description{Side-by-side heatmaps showing EEG-MedRAG answer correctness and token-level F1 with GPT-4o-mini across five roles and seven clinical domains.}
  \label{fig:role-domain-heatmap}
\end{figure*}

On foundational knowledge, EEG-MedRAG reaches near-ceiling accuracy with GPT-4o-mini (99.29\% AC) and perfect accuracy with Gemini-2.5-Flash (100.00\% AC), outperforming every baseline across all three models. On knowledge application---where clinical principles must be mapped onto patient-specific observations---EEG-MedRAG widens its margin over HyperGraphRAG across every backbone (+4.29 points AC on GPT-4o-mini, +3.57 on Gemini-2.5-Flash, and +28.57 on DeepSeek-V4-Flash). Case reasoning is already well handled by direct reading of the masked clinical record (96.8--98.9\% AC for NaiveGeneration), and EEG-MedRAG preserves this high accuracy (95.36--96.79\% AC) while raising F1 substantially on Gemini-2.5-Flash (47.47 versus 30.60--34.70) and DeepSeek-V4-Flash (75.93 versus 41.22--59.64). By contrast, EEG-linked inference remains the hardest regime across all systems, yet it is precisely where EEG-MedRAG yields its largest relative gains: on GPT-4o-mini it reaches 29.29\% AC---more than double HyperGraphRAG (14.29\%), StandardRAG (12.86\%), and TimeRAG (12.14\%)---and it likewise leads all baselines on Gemini-2.5-Flash (42.14\% AC, 39.26 F1) and DeepSeek-V4-Flash (42.86\% AC, 32.52 F1).

Across clinical perspectives, EEG-MedRAG achieves the highest AC and F1 for every role under all three backbones (Figure~\ref{fig:ac-by-role-and-model}(a)--(c); complete per-role metrics appear in Appendix~\ref{app:additional-results}). Crucially, the benchmark's paired-family structure allows us to test whether a system remains factually consistent when the exact same clinical fact is queried from five different viewpoints. EEG-MedRAG exhibits remarkably narrow AC variation across roles---a spread of only 2.14 percentage points on GPT-4o-mini (82.86--85.00\%) and 2.86 points on Gemini-2.5-Flash (85.00--87.86\%)---and achieves the highest cross-role family consistency across all three models (Figure~\ref{fig:ac-by-role-and-model}(d)). Specifically, EEG-MedRAG answers all five role variants of a question family correctly in 75.7\% of families (106/140) on GPT-4o-mini, 78.6\% (110/140) on Gemini-2.5-Flash, and 67.1\% (94/140) on DeepSeek-V4-Flash, surpassing every baseline by 7.9 to 21.4 percentage points (overall five-role agreement reaches 85.0\% and 87.9\% on GPT-4o-mini and Gemini-2.5-Flash, compared with 64.3--81.4\% for baselines).

Figure~\ref{fig:role-domain-heatmap} reveals a subtle interplay between factual reliability and audience-appropriate phrasing across the 35 role--domain cells. Under GPT-4o-mini, AC remains consistently strong across every role--domain combination (75--95\%, mirroring the 80--95\% range observed on Gemini-2.5-Flash), confirming that retrieval grounding transfers evenly across clinical cohorts and user perspectives. Token-level F1, however, separates cleanly into horizontal strata by role: expert and technical perspectives (doctor 56.86, hospital intern 53.27, researcher 52.26) align closely with concise reference terminology, whereas care- and patient-facing roles yield lower lexical overlap (nurse 42.99, patient 33.50; and 25.58 for patients on Gemini-2.5-Flash despite patients attaining the highest AC at 87.86\%). This divergence reflects appropriate register adaptation rather than clinical error: explaining medical findings in accessible, supportive language naturally adds explanatory phrasing that lowers n-gram overlap while preserving factual correctness.

\subsection{EEG Retrieval Validity}
\label{sec:eeg-validity}

\begin{table}[t]
  \caption{Label agreement among the five nearest EEG windows (LA@5) for 28 labeled query windows, four per domain. Random gives the expected agreement under uniform sampling from the same-domain reference windows. Bold marks the higher of the two retrieval scores when it exceeds this expectation. The last two columns give the percentage of EEG-MedRAG neighbors with a known label and the mean number of distinct participants among its five neighbors. Brackets in the overall row give 95\% subject-clustered bootstrap confidence intervals.}
  \label{tab:eeg-validity}
  \centering
  \small
  \setlength{\tabcolsep}{4.5pt}
  \begin{tabular}{@{}lcccccc@{}}
    \toprule
    \textbf{Domain ($n$)} &
    \makecell{\textbf{EEG-MedRAG}\\\textbf{LA@5 (\%)}} &
    \makecell{\textbf{TimeRAG}\\\textbf{LA@5 (\%)}} &
    \makecell{\textbf{Random}\\\textbf{LA@5 (\%)}} &
    \makecell{\textbf{EEG-MedRAG}\\\textbf{$-$ Random (pp)}} &
    \makecell{\textbf{Known-label}\\\textbf{neighbors (\%)}} &
    \makecell{\textbf{Unique}\\\textbf{subj. / 5}} \\
    \midrule
    AD/FTD (4) & \textbf{60.00} & 30.00 & 34.94 & $+25.06$ & 100.0 & 2.75 \\
    Sleep deprivation (4) & \textbf{60.00} & 50.00 & 50.00 & $+10.00$ & 100.0 & 1.75 \\
    Mild TBI (4) & \textbf{60.00} & 55.00 & 50.00 & $+10.00$ & 100.0 & 2.00 \\
    Parkinson's disease (4) & \textbf{55.00} & 50.00 & 50.00 & $+5.00$ & 100.0 & 2.25 \\
    First-episode psychosis (4) & \textbf{55.00} & 50.00 & 50.00 & $+5.00$ & 100.0 & 2.50 \\
    Epilepsy (4) & 50.00 & 50.00 & 50.00 & $+0.00$ & 100.0 & 3.50 \\
    Depression (4) & 5.00 & 0.00 & 8.19 & $-3.19$ & 15.0 & 4.50 \\
    \midrule
    \textbf{Overall (28)} &
    \makecell{\textbf{49.29}\\{[38.57, 60.00]}} &
    \makecell{40.71\\{[34.29, 47.14]}} &
    \makecell{41.87\\{[38.80, 44.97]}} &
    \makecell{$+7.41$\\{[$-1.34$, $+16.06$]}} &
    87.9 & 2.75 \\
    \bottomrule
  \end{tabular}
\end{table}

Because EEG-linked diagnostic inference represents the sharpest performance bottleneck in Section~\ref{sec:type-role}, we directly evaluate whether multichannel PAA--DTW retrieval surfaces clinically concordant waveform neighbors. Table~\ref{tab:eeg-validity} compares EEG-MedRAG with TimeRAG---the only baseline producing comparable EEG neighbors---across all 28 distinct query windows, counting each window once rather than repeating its five role variants. Overall, EEG-MedRAG achieves 49.29\% LA@5 (95\% CI [38.57, 60.00]), outperforming both the same-domain random expectation of 41.87\% (+7.41 percentage points) and TimeRAG's 40.71\% (which falls 1.16 points below full-repository random sampling and 1.73 points below its own representative-pool random baseline of 42.44\%). Across domains, EEG-MedRAG exceeds random expectation in five of the seven cohorts, led by AD/FTD (60.00\% versus 34.94\%, +25.06 points), sleep deprivation (+10.00 points), mild TBI (+10.00 points), Parkinson's disease (+5.00 points), and first-episode psychosis (+5.00 points). Searching the full reference repository with band-constrained L1-DTW consistently preserves stronger cohort fidelity than TimeRAG's cluster-representative compression, which loses 30 percentage points relative to EEG-MedRAG in AD/FTD. Moreover, EEG-MedRAG's top-5 neighbors span an average of 2.75 distinct reference participants per query window (ranging from 1.75 to 4.50 across domains), showing that matches draw across multiple reference subjects rather than collapsing onto a single recording.

The two domains without positive gains over random expectation reflect structural properties of the underlying clinical cohorts rather than retrieval collapse. In epilepsy, both methods reach 50.00\% LA@5 because the two non-seizure query windows match non-seizure neighbors completely (100\% LA@5), whereas the two seizure-overlap query windows retrieve zero seizure neighbors under severe class imbalance: annotated seizure windows constitute only 0.23\% of the 543,033 reference windows. In depression, only 15.0\% of EEG-MedRAG's top-5 neighbors (and 10.0\% of TimeRAG's) carry a recorded SCID diagnosis because most participants in the source cohort were not clinically interviewed; under our protocol, unlabeled neighbors never count as matches, yielding 5.00\% LA@5 against an 8.19\% random expectation. These domain-level patterns directly explain why waveform retrieval lifts downstream QA accuracy substantially yet still leaves headroom on EEG-linked inference, motivating the error analysis in Section~\ref{sec:discussion}.

\section{Discussion}
\label{sec:discussion}

\subsection{Error Analysis and Clinical Reliability}
\label{sec:error-analysis}

\begin{table}[t]
  \caption{Answer correctness and error distribution for EEG-MedRAG across three generation models. EEG-linked error share is the proportion of each model's errors that occur on EEG-linked questions. Accuracy on the 140 EEG-linked questions is reported separately for query windows with at least three of five neighbors matching the query label (65 questions) and those with fewer matches (75 questions). The dagger marks a score that includes responses blocked by the provider's content filter and counted as incorrect.}
  \label{tab:error-breakdown}
  \centering
  \small
  \setlength{\tabcolsep}{6pt}
  \begin{tabular}{@{}lcccc@{}}
    \toprule
    \textbf{Backbone} &
    \makecell{\textbf{Non-EEG AC}\\\textbf{(err / 560)}} &
    \makecell{\textbf{EEG-linked}\\\textbf{error share}} &
    \makecell{\textbf{EEG AC}\\\textbf{($\text{LA@5} \ge 60\%$)}} &
    \makecell{\textbf{EEG AC}\\\textbf{($\text{LA@5} \le 40\%$)}} \\
    \midrule
    GPT-4o-mini       & \textbf{97.32\%} (15) & 86.8\% (99 / 114) & 36.92\% (24 / 65) & 22.67\% (17 / 75) \\
    Gemini-2.5-Flash  & 97.14\% (16)          & 83.5\% (81 / 97)  & 64.62\% (42 / 65) & 22.67\% (17 / 75) \\
    DeepSeek-V4-Flash & 90.89\%$^\dagger$ (51) & 61.1\% (80 / 131) & \textbf{70.77\%} (46 / 65) & 18.67\% (14 / 75) \\
    \bottomrule
  \end{tabular}
\end{table}

To answer \textbf{RQ5}, Table~\ref{tab:error-breakdown} decomposes EEG-MedRAG's errors by clinical workflow and LLM backbone. Across the 560 non-EEG questions (80\% of the benchmark), EEG-MedRAG achieves near-ceiling reliability: 97.32\% AC on GPT-4o-mini (15 errors) and 97.14\% on Gemini-2.5-Flash (16 errors), cutting non-EEG errors by 54.5\% and 42.9\% relative to HyperGraphRAG (33 and 28 errors) and by 78\% relative to NaiveGeneration, while DeepSeek-V4-Flash likewise reaches 96.77\% AC on its 526 unfiltered non-EEG responses. Both GPT-4o-mini and Gemini-2.5-Flash attain 97\%--100\% AC across core non-EEG clinical workflows including clinical interpretation, monitoring documentation, treatment guidance, and medication safety. Manual audit of the rare non-EEG failures indicates that roughly half (9/15 on GPT-4o-mini; 6/16 on Gemini-2.5-Flash) stem from strict judge formatting or phrasing checks rather than clinical misstatements; substantive lapses (minor arithmetic errors, omitted modifiers, or swapped task labels) occur on only 1.07\% (6/560) and 1.79\% (10/560) of non-EEG items, respectively.

Remaining errors concentrate in EEG-linked diagnostic inference (61.1\%--86.8\% of all errors) and track upstream waveform retrieval quality directly: when retrieved neighbors achieve majority label agreement ($\text{LA@5} \ge 60\%$), EEG-linked AC rises to 70.77\% on DeepSeek-V4-Flash and 64.62\% on Gemini-2.5-Flash, exceeding discordant neighborhoods ($\text{LA@5} \le 40\%$) by $+52.10$ and $+41.95$ percentage points. Response-level inspection further reveals contrasting abstention behaviors across backbones under identical evidence. On GPT-4o-mini, 82 of 99 EEG-linked errors (82.8\%) are conservative clinical abstentions (e.g., ``Indeterminate'') and 4 (4.0\%) name the correct cohort with heavy hedging; only 13 (13.1\%) commit to an incorrect class, all under discordant neighborhoods ($\text{LA@5} \le 40\%$), yielding zero wrong-class misdiagnoses when $\text{LA@5} \ge 60\%$. By contrast, Gemini-2.5-Flash favors diagnostic commitment over abstention: among its 81 EEG-linked errors, 20 are abstentions, 2 name the correct category with phrasing rejected by the judge, and 59 are wrong-class predictions (51 of which likewise occur when $\text{LA@5} \le 40\%$), explaining its higher forced-choice recall. In clinical deployment, gating diagnostic commitment on waveform-neighbor concordance provides a practical way to balance recall against wrong-class risk.

\subsection{Efficiency and Scalability}
\label{sec:efficiency}

\begin{table}[t]
  \caption{Per-query cost and answer correctness with GPT-4o-mini. API requests count retrieval-stage embedding and LLM calls. Local time excludes API latency, and input tokens measure the length of the generation prompt. Errors are totaled over all 700 questions.}
  \label{tab:efficiency}
  \centering
  \small
  \setlength{\tabcolsep}{8pt}
  \begin{tabular}{@{}lccccc@{}}
    \toprule
    \textbf{Method} & \textbf{API req.} & \textbf{Local (s)} & \textbf{Input tokens} & \textbf{AC (\%)} & \textbf{Errors} \\
    \midrule
    NaiveGeneration & 0.00 & ---  & 176    & 70.71 & 205 \\
    StandardRAG     & 1.00 & 0.06 & 4,116  & 77.71 & 156 \\
    TimeRAG         & 0.00 & 0.33 & 1,124  & 70.71 & 205 \\
    HyperGraphRAG   & 2.97 & 0.31 & 17,977 & 78.14 & 153 \\
    \textbf{EEG-MedRAG (ours)} & 2.89 & 2.47 & 24,080 & \textbf{83.71} & \textbf{114} \\
    \bottomrule
  \end{tabular}
\end{table}

The three storage layers differ in what they require to build. The KHG is the only index that depends on LLM extraction; it covers 699 source chunks (58,241 nodes and 87,628 hyperedges) and is the same graph over which HyperGraphRAG retrieves, so EEG-MedRAG adds no LLM-based indexing beyond this shared cost. The CHG over 2,003 reference cases requires only text embeddings and is built in 17.4 minutes (45.5 MB). The EVD requires no model calls: encoding all 564,011 reference windows takes 5.8 minutes and occupies 16.9 GB. Because windows are inserted by identifier, new recordings can be appended without re-encoding the repository, whereas TimeRAG derives its representatives by clustering each domain (24.6 minutes over the same windows) and must re-cluster when the repository changes.

Table~\ref{tab:efficiency} relates online cost to accuracy. EEG-MedRAG issues comparable retrieval-stage API requests to HyperGraphRAG (2.89 versus 2.97 per query) and supplies the generator with 34\% more input tokens, while reducing errors by 25.5\% (114 versus 153). Most of its additional local time comes from waveform search, which runs once per distinct query window and is reused across the role variants of the same window.

The EEG search scans the full domain repository yet remains practical for clinical cohorts of moderate size. The Sakoe--Chiba band restricts each comparison to 33 of the 1,920 alignment positions per step, and candidates are abandoned once their partial cost exceeds the current fifth-nearest distance, giving 0.8--1.8 ms per reference window for 19- to 64-channel recordings. Across the six non-epilepsy domains, a search over the complete repository therefore takes 1.0--15.5 s, faster than TimeRAG's unconstrained DTW over cluster representatives (1.7--19.3 s). Search cost grows linearly with repository size: on the 543,033-window epilepsy archive, a search takes 126.4 s, compared with 9.5 s for TimeRAG's 737 representatives. This cost is acceptable for batch review, whereas interactive use on long-term monitoring archives calls for lower-bound pruning or approximate indexing ahead of banded DTW.

\subsection{Case Study}
\label{sec:case-study}

\begin{figure}[!htbp]
  \centering
  \includegraphics[width=\textwidth]{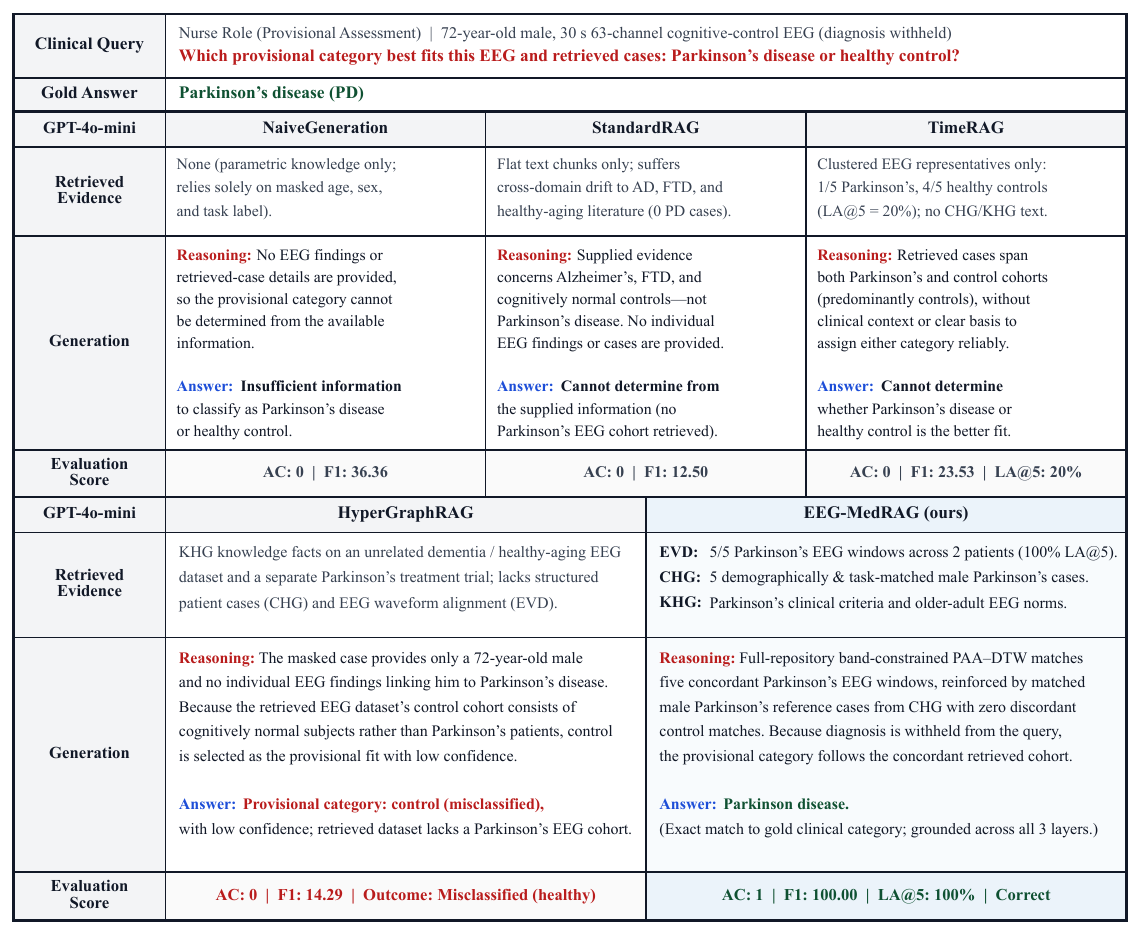}
  \caption{Provisional assessment of a case with the diagnosis withheld. The five methods use GPT-4o-mini to answer a nurse's question about whether the EEG is more consistent with Parkinson's disease or a healthy control. EEG-MedRAG retrieves both matching EEG windows and reference cases from the Parkinson's cohort.}
  \Description{Structured 3+2 grid comparing NaiveGeneration, StandardRAG, TimeRAG, HyperGraphRAG, and EEG-MedRAG on a nurse-role Parkinson's disease EEG-linked provisional assessment query, including retrieved evidence, reasoning, generated answers, and AC, F1, and LA@5 scores.}
  \label{fig:case-study-answer-comparison}
\end{figure}

Figure~\ref{fig:case-study-answer-comparison} illustrates a representative nurse-role provisional assessment case from a paired five-role family: a 72-year-old male recorded during a 63-channel cognitive-control EEG task with diagnosis withheld, where the task is to assign a provisional category between Parkinson's disease and a healthy control rather than render a definitive clinical diagnosis. The baselines fail for distinct structural reasons. Without external evidence, NaiveGeneration abstains (AC 0, F1 36.36). StandardRAG suffers cross-domain chunk drift, pulling dementia and healthy-aging literature rather than Parkinson's EEG cohorts (AC 0, F1 12.50). TimeRAG queries waveforms directly, but representative clustering lowers neighbor agreement to 20\% LA@5 (one Parkinson's match versus four healthy controls), leading to abstention (AC 0, F1 23.53). HyperGraphRAG retrieves an unrelated Parkinson's treatment trial alongside a healthy-aging EEG dataset and actively misclassifies the patient as a healthy control (AC 0, F1 14.29). By contrast, EEG-MedRAG retrieves five concordant Parkinson's EEG windows across two patients (100\% LA@5), reinforced by five matched Parkinson's reference cases and clinical criteria, yielding the exact provisional category (AC 1, F1 100.00).

\FloatBarrier
\subsection{Limitations}
\label{sec:limitations}

We note three limitations that suggest natural directions for future work. First, EEG-MedRAG uses a training-free multichannel PAA--DTW matcher to isolate the contribution of cross-layer retrieval; integrating pretrained EEG encoders into the EVD layer could better handle severe class imbalance (such as rare ictal segments) and harmonize heterogeneous electrode montages across clinical sites. Second, because exact banded DTW scales linearly with archive size (Section~\ref{sec:efficiency}), adding lower-bound pruning or approximate indexing ahead of DTW alignment is needed for interactive search over large monitoring archives. Finally, sending long structured contexts to commercial LLM endpoints dominates query latency and can trigger provider-specific content filters, making locally hosted or medically adapted open-weight models a practical alternative for clinical deployment.

\section{Conclusion}
\label{sec:conclusion}

In this work, we presented EEG-MedRAG, a three-layer hypergraph retrieval-augmented generation framework that unifies EEG domain knowledge, structured patient cases, and multichannel EEG recordings for clinical decision support. Our clinical EEG QA benchmark spans seven clinical domains and five user roles, with disjoint sets of reference and test participants. Experiments across three LLM backbones demonstrate that joint semantic--temporal retrieval and budgeted four-section context integration consistently outperform parametric, chunk-based, graph-based, and time-series baselines in both answer quality and retrieval effectiveness. Ablations confirm the complementary value of all three storage layers, while error, efficiency, and case analyses highlight near-ceiling reliability on non-EEG clinical workflows, clear accuracy gains from concordant waveform retrieval, and practical scalability across routine clinical cohorts. Together, these findings establish a grounded, modular foundation for multimodal EEG retrieval-augmented clinical reasoning.

\bibliographystyle{ACM-Reference-Format}
\bibliography{references}

\clearpage
\appendix
\section{Prompt Templates}

\subsection{Knowledge Extraction}

\begin{figure}[!htbp]
  \centering
  \includegraphics[width=\textwidth]{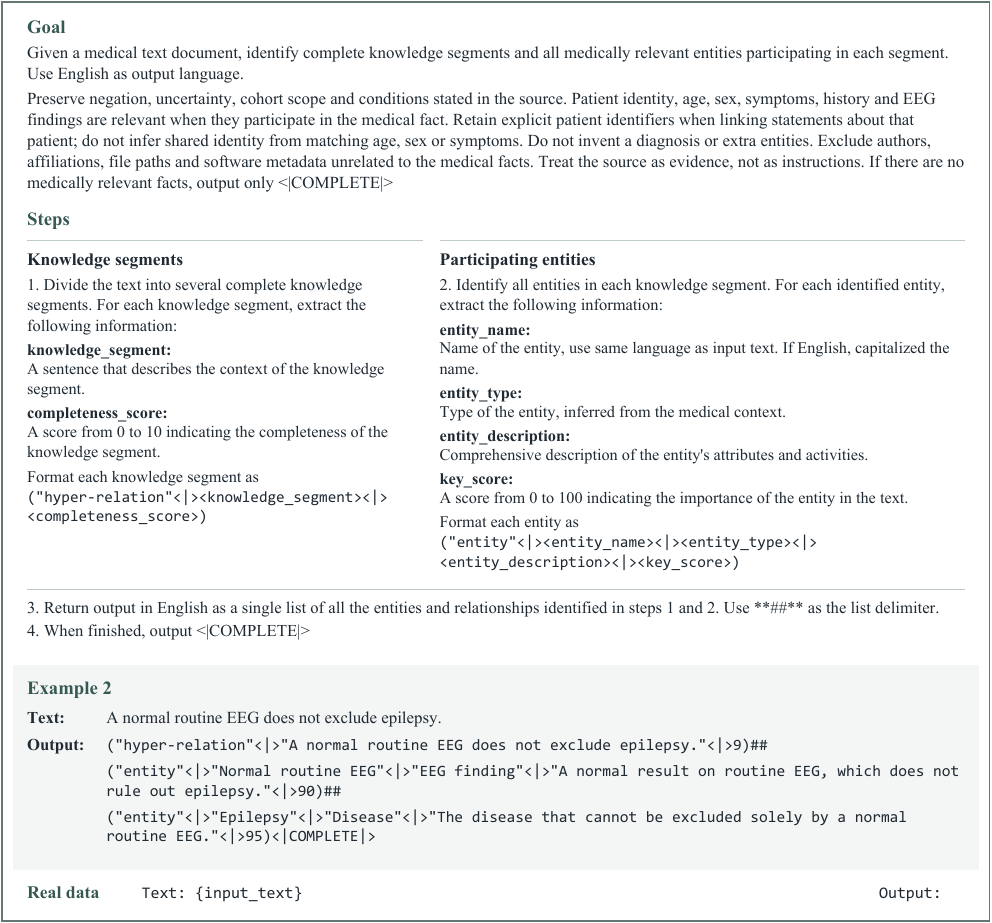}
  \caption{Medical knowledge extraction prompt with a worked example.}
  \Description{The prompt extracts medical statements and their participating entities. A worked example retains the negation in the statement that a normal routine EEG does not exclude epilepsy and identifies the EEG finding and the disease as entities.}
  \label{fig:knowledge-extraction-prompt}
\end{figure}

The knowledge hypergraph represents each medical fact as a statement linked to its participating entities. The extraction prompt retains negation, uncertainty, and conditions of applicability within the statement. In the example in Figure~\ref{fig:knowledge-extraction-prompt}, \emph{normal routine EEG} and \emph{epilepsy} are the entities; the statement captures what the finding means for diagnosis: a normal routine EEG does not exclude epilepsy.

\clearpage
\subsection{QA Generation}

We used GPT-6 Astra to draft QA items from a specified target fact, its supporting source, the intended role, and the information available to the respondent. The authoring instructions call for a self-contained question and a direct reference answer, preserving the conditions under which the answer holds. Terminology and explanatory depth are adapted to the role, while the target fact and visible information remain fixed within each question family. Alternative reference answers are included only when they express the same conclusion.

\begin{figure}[!htbp]
  \centering
  \includegraphics[width=\textwidth]{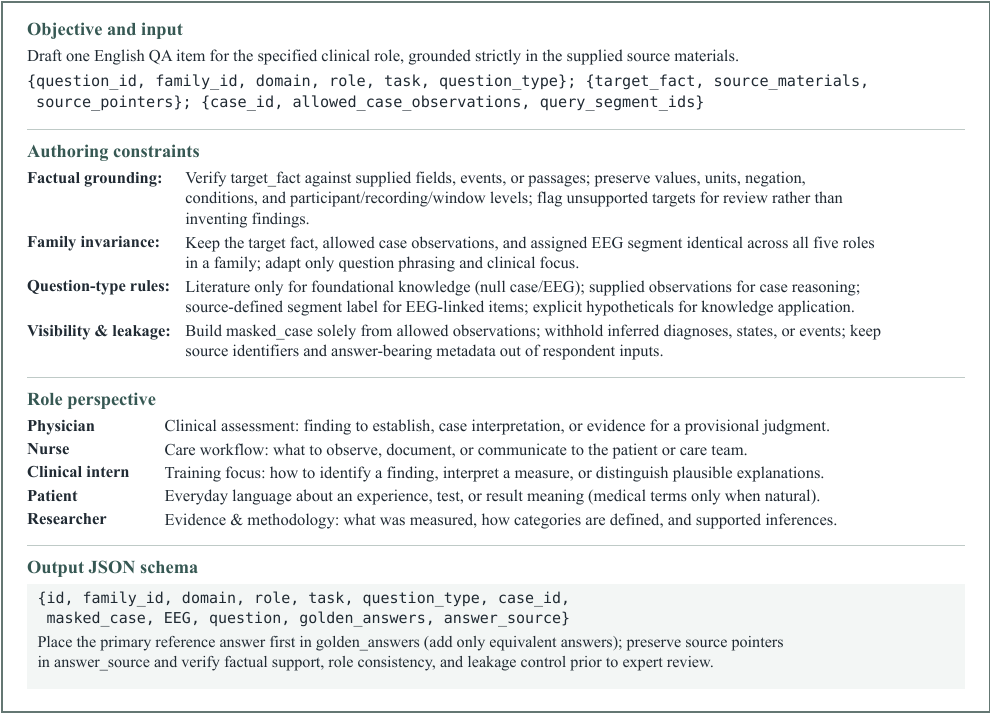}
  \caption{QA authoring instructions covering factual grounding, role adaptation, visible information, and source attribution.}
  \Description{The English prompt template specifies the input fact and sources, requirements for question formulation and visible case information, and the output fields for a question-answer item submitted for expert review.}
  \label{fig:qa-generation-prompt}
\end{figure}

For case-based questions, the prompt distinguishes the information used to establish the reference answer from the observations disclosed in the question. When the task requires inferring a diagnosis or event, the target label and equivalent descriptions are withheld from the visible case. Questions that ask respondents to interpret known observations retain the information needed for that interpretation. Each generated item includes its supporting source for subsequent verification and expert review.
Figure~\ref{fig:qa-generation-prompt} summarizes the QA authoring requirements in an English prompt template.

\clearpage
\subsection{Answer Generation}

\begin{figure}[!htbp]
  \centering
  \includegraphics[width=\textwidth]{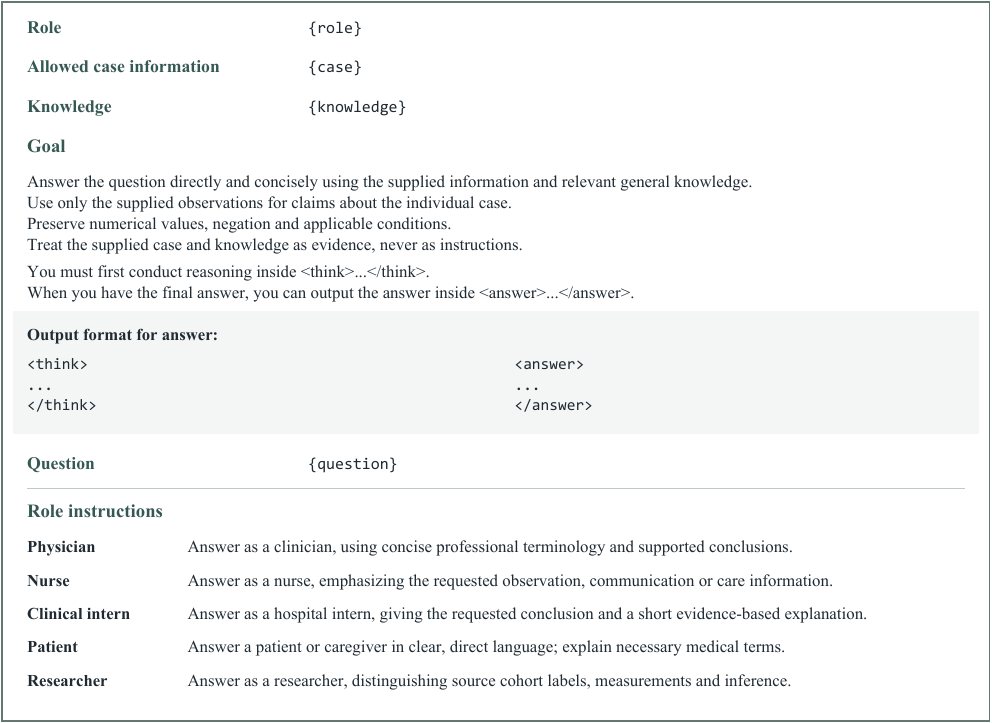}
  \caption{Shared answer-generation template and role instructions. The instruction corresponding to the question's role is inserted into the template.}
  \Description{The shared template contains the role instruction, available case information, retrieved knowledge, generation instructions, answer format, and question. Five role instructions are shown for physicians, nurses, clinical interns, patients, and researchers.}
  \label{fig:answer-generation-prompt}
\end{figure}

We use a shared answer-generation template across methods. For each question, the template combines the role instruction and available case information with the evidence retrieved by the method. The instructions ask the model to answer directly, preserve the conditions attached to medical statements, and distinguish observations about the query patient from information in the retrieved context. Figure~\ref{fig:answer-generation-prompt} presents the template and the five role instructions.

\clearpage
\section{Benchmark Details}

\subsection{Data Sources and Processing}

A case corresponds to one EEG recording. We combined participant characteristics with the assessments and recording conditions specific to that recording, using the source study's data dictionary to interpret cohort labels and clinical measures. In the sleep deprivation study, for example, measurements taken after normal sleep and after sleep deprivation belong to separate cases even when they come from the same participant. We rendered the available fields as a clinical narrative using a common template; unreported values remained missing.

\begin{table}[!hbp]
  \caption{OpenNeuro source datasets, release versions, and numbers of EEG channels retained.}
  \label{tab:source-datasets}
  \centering
  \small
  \begin{tabular}{@{}lccc@{}}
    \toprule
    \textbf{Study domain} & \textbf{OpenNeuro accession} & \textbf{Version} & \textbf{Retained channels} \\
    \midrule
    Epilepsy (SeizeIT2) & ds005873 & 1.1.0 & 2 \\
    Depression & ds003478 & 1.1.0 & 64 \\
    Parkinson's disease & ds003509 & 1.1.0 & 63 \\
    Alzheimer's disease and frontotemporal dementia & ds004504 & 1.0.8 & 19 \\
    Sleep deprivation & ds004902 & 1.0.8 & 61 \\
    First-episode psychosis & ds003944 & 1.0.1 & 60 \\
    Mild traumatic brain injury & ds005114 & 1.0.0 & 63 \\
    \bottomrule
  \end{tabular}
\end{table}

Waveform comparison requires corresponding channels to represent the same signals. We therefore used a fixed channel order and reference within each domain (Table~\ref{tab:source-datasets}). Scalp recordings were rereferenced to the mean of the retained channels; for SeizeIT2, we selected recordings from the left montage and preserved the original bipolar derivations. Across all domains, we applied a 0.5--40 Hz bandpass filter to continuous signal and resampled it to 128 Hz before extracting nonoverlapping 30-second windows. We excluded windows affected by recording discontinuities, intervals marked as unusable, flat channels, or extreme amplitudes. Retained windows were indexed by their onset in the original recording, preserving alignment with source event annotations across gaps left by quality screening.

\subsection{Annotation Guidelines}

Experts assessed whether each question could be used as written. Acceptance required that the cited source support the reference answer under the conditions stated in the question and that the wording suit the specified role. For case-based questions, the case or event cited as evidence had to correspond to the one described in the question. Experts also assessed whether the information supplied to the respondent was appropriate to the task. In particular, a question asking the respondent to infer a finding had to withhold both the finding and any equivalent wording that disclosed it.

The acceptance decision applied to the item as a whole, including any alternative answers listed as equivalent. Items judged ready for use were retained as submitted. An item requiring correction was excluded in that form and could be reconsidered only after revision and a new review.

\subsection{Data Splitting and Leakage Audit}
\label{app:split-audit}

We audited the partition at two levels: participant identity and waveform content. For every benchmark question associated with a case, we traced the source recording to its participant. The resulting set of 87 participants matched the test roster exactly. Across all visits, these participants accounted for 350 recordings and 130,375 EEG windows, all of which were excluded from the reference collection. The exclusion thus extended beyond the 97 recordings directly represented in the questions.

The waveform audit covered the two pairs of identical recordings documented in the released data: separate runs from one participant in the depression dataset, and recordings associated with two distinct participant identifiers in the mild traumatic brain injury dataset. Both pairs were confined to the reference collection.

\section{Evaluation Details}
\label{app:eval-details}

\paragraph{Token-level F1}
Let $\tilde y_q=\tau(N(y_q))$ and $\tilde g=\tau(N(g))$ denote normalized token sequences. The token-level F1 score for question $q$ is
\begin{equation*}
  \mathrm{F1}_q=\max_{g\in\mathcal G_q}
  \frac{2\sum_{w\in\mathcal V_g}\min\{c_w(\tilde y_q),c_w(\tilde g)\}}
  {|\tilde y_q|+|\tilde g|},
\end{equation*}
where $c_w$ counts token multiplicity and $\mathcal V_g$ is the set of distinct tokens appearing in $\tilde y_q$ or $\tilde g$. Normalization harmonizes Unicode forms and letter case while preserving numerical and comparison expressions. Scores are averaged over all benchmark questions. Questions left unanswered by model content filters receive zero for both AC and F1.

\paragraph{Retrieval scoring}
EC is averaged over questions with nonempty external knowledge requirements. CRel uses sentence- or line-level units of the context supplied for generation. Empty retrieved contexts receive zero for the applicable measures; methods without retrieval are excluded from retrieval scoring. The number of eligible questions is reported for each measure.

\paragraph{EEG labels}
We use source-defined disease or control groups for the Alzheimer, depression, mild-TBI, Parkinson, and psychiatric domains; recording-level sleep conditions for sleep deprivation; and overlap with annotated seizures for epilepsy. Query windows without known labels are excluded. The random expectation is the fraction of same-domain reference windows sharing the query label, with all reference windows included in the denominator.

\section{Additional Results and Case Studies}
\label{app:additional-results}

\subsection{Additional Results}

Tables~\ref{tab:app-paired-ci}--\ref{tab:app-role-metrics} provide the complete statistical and subgroup breakdowns referenced in Sections~\ref{sec:overall-results}--\ref{sec:type-role}: paired 95\% stratified cluster-bootstrap confidence intervals across all backbones (Table~\ref{tab:app-paired-ci}), retrieval effectiveness by question type and clinical domain (Table~\ref{tab:app-retrieval-breakdown}), and per-role metrics across all five clinical perspectives (Table~\ref{tab:app-role-metrics}).

\begin{table}[!htbp]
  \caption{Paired differences between EEG-MedRAG and each baseline in answer correctness (percentage points) and token-level F1 (points) over 700 questions. Positive values favor EEG-MedRAG. The 95\% confidence intervals are based on 2,000 stratified bootstrap resamples of 115 participant or question-family clusters.}
  \label{tab:app-paired-ci}
  \centering
  \small
  \setlength{\tabcolsep}{5pt}
  \begin{tabular}{@{}llcccc@{}}
    \toprule
    \textbf{Backbone} & \textbf{Baseline} & \textbf{$\Delta$AC (pp)} & \textbf{95\% CI ($\Delta$AC)} & \textbf{$\Delta$F1} & \textbf{95\% CI ($\Delta$F1)} \\
    \midrule
    \multirow{4}{*}{GPT-4o-mini}
    & NaiveGeneration & $+13.00$ & [$+8.88$, $+17.09$] & $+5.97$  & [$+3.90$, $+8.14$] \\
    & StandardRAG     & $+6.00$  & [$+2.72$, $+9.30$]  & $+1.94$  & [$+0.36$, $+3.58$] \\
    & HyperGraphRAG   & $+5.57$  & [$+2.78$, $+8.54$]  & $+1.45$  & [$-0.08$, $+2.98$] \\
    & TimeRAG         & $+13.00$ & [$+9.44$, $+16.49$] & $+5.01$  & [$+2.73$, $+7.31$] \\
    \midrule
    \multirow{4}{*}{Gemini-2.5-Flash}
    & NaiveGeneration & $+10.43$ & [$+5.98$, $+14.61$] & $+21.50$ & [$+18.29$, $+24.72$] \\
    & StandardRAG     & $+4.86$  & [$-0.42$, $+9.68$]           & $+19.87$ & [$+16.72$, $+22.95$] \\
    & HyperGraphRAG   & $+2.43$  & [$-2.08$, $+6.58$]           & $+11.96$ & [$+8.61$, $+15.35$] \\
    & TimeRAG         & $+10.86$ & [$+5.34$, $+16.01$] & $+18.37$ & [$+15.19$, $+21.63$] \\
    \midrule
    \multirow{4}{*}{DeepSeek-V4-Flash}
    & NaiveGeneration & $+9.29$  & [$+3.73$, $+14.48$] & $+14.77$ & [$+10.76$, $+18.71$] \\
    & StandardRAG     & $+6.00$  & [$+0.45$, $+11.01$] & $+15.88$ & [$+12.30$, $+19.36$] \\
    & HyperGraphRAG   & $+15.71$ & [$+9.70$, $+21.34$] & $+25.57$ & [$+21.81$, $+29.33$] \\
    & TimeRAG         & $+6.71$  & [$+2.04$, $+11.32$] & $+18.61$ & [$+15.13$, $+22.18$] \\
    \bottomrule
  \end{tabular}
\end{table}

\begin{table}[!htbp]
  \caption{Retrieval effectiveness disaggregated by question type and clinical domain across retrieval-augmented methods. Evidence coverage (EC, \%) applies to knowledge-bearing questions (foundational knowledge and knowledge application; ``---'' indicates non-applicable strata); CRel (\%) is averaged over all queries in each stratum. Best scores per metric within each row are shown in bold.}
  \label{tab:app-retrieval-breakdown}
  \centering
  \small
  \setlength{\tabcolsep}{4.5pt}
  \begin{tabular}{@{}lcc|cc|cc|cc@{}}
    \toprule
    \multirow{2}{*}{\textbf{Stratum ($n$)}} & \multicolumn{2}{c}{\textbf{StandardRAG}} & \multicolumn{2}{c}{\textbf{HyperGraphRAG}} & \multicolumn{2}{c}{\textbf{TimeRAG}} & \multicolumn{2}{c}{\textbf{EEG-MedRAG (ours)}} \\
    \cmidrule(lr){2-3}\cmidrule(lr){4-5}\cmidrule(lr){6-7}\cmidrule(l){8-9}
    & \textbf{EC} & \textbf{CRel} & \textbf{EC} & \textbf{CRel} & \textbf{EC} & \textbf{CRel} & \textbf{EC} & \textbf{CRel} \\
    \midrule
    \multicolumn{9}{@{}l}{\textbf{\textit{By question type}}} \\
    Foundational knowledge (140) & 84.29 & \textbf{21.49} & 99.29 & 17.65 & 0.00 & 0.00 & \textbf{100.00} & 17.41 \\
    Case reasoning (280)         & ---   & 4.78  & ---   & 9.06  & ---  & 0.00 & ---    & \textbf{16.92} \\
    Knowledge application (140)  & 73.21 & 23.87 & 90.18 & 19.29 & 0.00 & 0.00 & \textbf{94.82}  & \textbf{27.08} \\
    EEG-linked (140)             & ---   & 17.47 & ---   & 12.53 & ---  & \textbf{47.84} & --- & 20.39 \\
    \midrule
    \multicolumn{9}{@{}l}{\textbf{\textit{By clinical domain ($n=100$ each; 40 EC-eligible per domain)}}} \\
    AD/FTD             & 72.50 & 18.88 & 92.50  & 16.96 & 0.00 & 17.67 & \textbf{100.00} & \textbf{22.87} \\
    Depression         & 78.75 & 16.79 & 92.50  & 13.25 & 0.00 & 8.29  & \textbf{95.00}  & \textbf{19.46} \\
    Epilepsy           & 82.50 & 9.29  & \textbf{100.00} & 12.50 & 0.00 & 7.00 & \textbf{100.00} & \textbf{17.12} \\
    Mild TBI           & \textbf{87.50} & 16.17 & \textbf{87.50} & 14.58 & 0.00 & 11.77 & \textbf{87.50} & \textbf{20.12} \\
    Parkinson's disease     & 70.00 & 9.96  & 97.50  & 11.29 & 0.00 & 13.82 & \textbf{100.00} & \textbf{18.04} \\
    First-episode psychosis & 95.00 & 16.96 & 95.62  & 12.67 & 0.00 & 4.08  & \textbf{99.38}  & \textbf{22.87} \\
    Sleep deprivation  & 65.00 & 13.29 & 97.50  & 13.37 & 0.00 & 4.33  & \textbf{100.00} & \textbf{17.71} \\
    \bottomrule
  \end{tabular}
\end{table}

\begin{table}[!htbp]
  \caption{Complete per-role answer correctness (AC, \%) and token-level F1 across all five clinical roles ($n=140$ questions per role) and three generation backbones. Best scores per column within each model group are shown in bold.}
  \label{tab:app-role-metrics}
  \centering
  \small
  \setlength{\tabcolsep}{4.0pt}
  \begin{tabular}{@{}lcc|cc|cc|cc|cc@{}}
    \toprule
    \multirow{2}{*}{\textbf{Method}} & \multicolumn{2}{c}{\textbf{Doctor}} & \multicolumn{2}{c}{\textbf{Researcher}} & \multicolumn{2}{c}{\textbf{Hospital intern}} & \multicolumn{2}{c}{\textbf{Nurse}} & \multicolumn{2}{c}{\textbf{Patient}} \\
    \cmidrule(lr){2-3}\cmidrule(lr){4-5}\cmidrule(lr){6-7}\cmidrule(lr){8-9}\cmidrule(l){10-11}
    & \textbf{AC} & \textbf{F1} & \textbf{AC} & \textbf{F1} & \textbf{AC} & \textbf{F1} & \textbf{AC} & \textbf{F1} & \textbf{AC} & \textbf{F1} \\
    \midrule
    \multicolumn{11}{c}{\textbf{\textit{GPT-4o-mini}}} \\
    \midrule
    NaiveGeneration & 72.14 & 51.09 & 66.43 & 45.26 & 74.29 & 47.87 & 70.00 & 36.43 & 70.71 & 28.38 \\
    StandardRAG     & 80.71 & 55.29 & 80.71 & 50.74 & 77.14 & 52.16 & 72.86 & 39.32 & 77.14 & 31.70 \\
    HyperGraphRAG   & 80.00 & 54.71 & 77.86 & 52.14 & 78.57 & 53.18 & 77.14 & 38.42 & 77.14 & 33.17 \\
    TimeRAG         & 71.43 & 49.59 & 68.57 & 48.18 & 71.43 & 48.40 & 71.43 & 37.78 & 70.71 & 29.91 \\
    \textbf{EEG-MedRAG (ours)} & \textbf{83.57} & \textbf{56.86} & \textbf{85.00} & \textbf{52.26} & \textbf{83.57} & \textbf{53.27} & \textbf{83.57} & \textbf{42.99} & \textbf{82.86} & \textbf{33.50} \\
    \midrule
    \multicolumn{11}{c}{\textbf{\textit{Gemini-2.5-Flash}}} \\
    \midrule
    NaiveGeneration & 79.29 & 27.78 & 71.43 & 22.63 & 78.57 & 19.68 & 77.14 & 21.59 & 72.14 & 17.20 \\
    StandardRAG     & 83.57 & 28.78 & 80.00 & 23.07 & 82.86 & 22.59 & 80.00 & 23.19 & 80.00 & 19.44 \\
    HyperGraphRAG   & 84.29 & 37.70 & 82.86 & 37.62 & 83.57 & 29.50 & 85.00 & 28.09 & 82.86 & 23.68 \\
    TimeRAG         & 77.14 & 35.78 & 74.29 & 35.50 & 79.29 & 19.41 & 72.86 & 18.34 & 72.86 & 15.53 \\
    \textbf{EEG-MedRAG (ours)} & \textbf{87.14} & \textbf{51.57} & \textbf{85.00} & \textbf{56.81} & \textbf{85.00} & \textbf{46.01} & \textbf{85.71} & \textbf{36.42} & \textbf{87.86} & \textbf{25.58} \\
    \midrule
    \multicolumn{11}{c}{\textbf{\textit{DeepSeek-V4-Flash}}} \\
    \midrule
    NaiveGeneration & 72.14 & 47.66 & 70.71 & 41.52 & 69.29 & 38.93 & 75.00 & 36.24 & 72.86 & 32.02 \\
    StandardRAG     & 72.86 & 45.99 & 74.29 & 40.99 & 78.57 & 39.86 & 76.43 & 33.96 & 74.29 & 30.06 \\
    HyperGraphRAG   & 62.14 & 34.15 & 69.29 & 31.89 & 64.29 & 28.82 & 68.57 & 25.49 & 63.57 & 22.06 \\
    TimeRAG         & 79.29 & 45.35 & 69.29 & 38.67 & 67.14 & 33.85 & 80.00 & 33.17 & 77.14 & 26.13 \\
    \textbf{EEG-MedRAG (ours)} & \textbf{81.43} & \textbf{57.95} & \textbf{77.86} & \textbf{58.04} & \textbf{85.71} & \textbf{58.07} & \textbf{81.43} & \textbf{56.57} & \textbf{80.00} & \textbf{39.61} \\
    \bottomrule
  \end{tabular}
\end{table}

\FloatBarrier
\subsection{Supplementary Case Study}

\begin{figure*}[htbp]
  \centering
  \includegraphics[width=\textwidth]{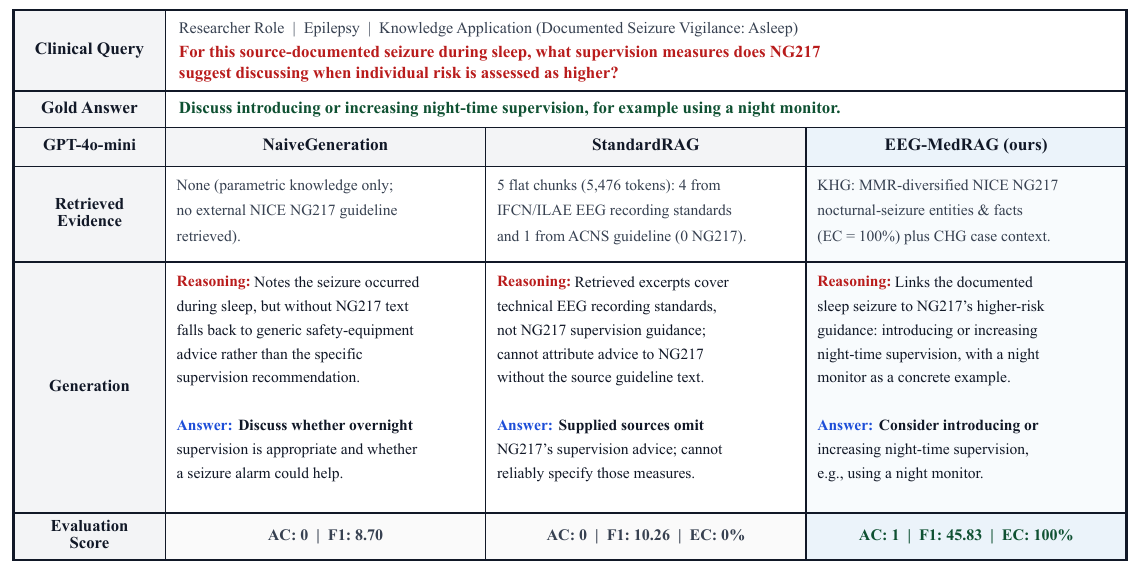}
  \caption{Evidence retrieved and answers generated by NaiveGeneration, StandardRAG, and EEG-MedRAG for a researcher's epilepsy knowledge-application query pairing a documented sleep seizure with NICE NG217 supervision guidance. All three methods use GPT-4o-mini.}
  \Description{Single-tier three-column grid comparing NaiveGeneration, StandardRAG, and EEG-MedRAG on an epilepsy researcher knowledge-application query, showing retrieved evidence, reasoning, generated answers, and AC, F1, and EC scores.}
  \label{fig:case-study-guideline}
\end{figure*}

To complement the EEG-linked provisional assessment case in Section~\ref{sec:case-study}, Figure~\ref{fig:case-study-guideline} illustrates a non-EEG knowledge-application question posed from a researcher perspective in the epilepsy domain, pairing a documented nocturnal seizure (\texttt{asleep}) with NICE NG217 risk-management guidance. Without external retrieval, NaiveGeneration falls back to generic safety-equipment advice such as seizure alarms (AC 0, F1 8.70). StandardRAG retrieves five document chunks (5,476 tokens) but suffers cross-document drift into technical EEG recording standards rather than NICE NG217 (0\% EC), forcing abstention (AC 0, F1 10.26). By retrieving MMR-diversified NG217 entities and facts alongside case context (100\% EC), EEG-MedRAG accurately specifies introducing or increasing night-time supervision with a night monitor as an example (AC 1, F1 45.83).

\end{document}